\documentclass{article}%
\usepackage{amsmath}
\usepackage{amsfonts}
\usepackage{amssymb}
\usepackage{graphicx}
\usepackage{algorithmic}
\usepackage{algorithm}%
\providecommand{\U}[1]{\protect\rule{.1in}{.1in}}
\begin{document}

\title{SurvNAM: The machine learning survival model explanation}
\author{Lev V. Utkin, Egor D. Satyukov and Andrei V. Konstantinov\\Peter the Great St.Petersburg Polytechnic University\\St.Petersburg, Russia\\e-mail: lev.utkin@gmail.com, satukov-egor@mail.ru, andrue.konst@gmail.com}
\date{}
\maketitle

\begin{abstract}
A new modification of the Neural Additive Model (NAM) called SurvNAM and its
modifications are proposed to explain predictions of the black-box machine
learning survival model. The method is based on applying the original NAM to
solving the explanation problem in the framework of survival analysis. The
basic idea behind SurvNAM is to train the network by means of a specific
expected loss function which takes into account peculiarities of the survival
model predictions and is based on approximating the black-box model by the
extension of the Cox proportional hazards model which uses the well-known
Generalized Additive Model (GAM) in place of the simple linear relationship of
covariates. The proposed method SurvNAM allows performing the local and global
explanation. A set of examples around the explained example is randomly
generated for the local explanation. The global explanation uses the whole
training dataset. The proposed modifications of SurvNAM are based on using the
Lasso-based regularization for functions from GAM and for a special
representation of the GAM functions using their weighted linear and non-linear
parts, which is implemented as a shortcut connection. A lot of numerical
experiments illustrate the SurvNAM efficiency.


\end{abstract}

\section{Introduction}

An increasing importance of machine learning models, especially deep learning
models, and their spreading incorporation into many applied problems lead to
the problem of the prediction explanation and interpretation. The problem
stems from the fact that many powerful and efficient machine learning models
are viewed as black boxes, i.e., details of their functioning are often
completely unknown. This implies that the model predictions cannot be easily
explained because users of the models do not know how predictions have been
achieved. A vivid example is the patient diagnosis prediction in medicine when
a doctor needs clear understanding of the stated diagnosis, understanding why
the machine learning model predicts a certain disease. This understanding may
help in choosing a desirable treatment \cite{Holzinger-etal-2019}. The
importance of the explanation problem leads to developing many methods and
models which try to explain and to interpret predictions of the machine
learning models
\cite{Arya-etal-2019,Belle-Papantonis-2020,Guidotti-2019,Liang-etal-2021,Molnar-2019,Murdoch-etal-2019,Xie-Ras-etal-2020,Zablocki-etal-21,Zhang-Tino-etal-2020}%
.

Explanation methods can be divided into two sets: local and global methods.
Local methods try to explain a black-box model locally around a test example.
In contrast to local methods, the global ones try to explain predictions for
the whole dataset. In spite of importance and interest of the global
interpretation methods, most applications aim to understand decisions
concerning with an object to answer the question what features of the analyzed
object are responsible for a black-box model prediction. Therefore, we focus
on both the groups of interpretation methods, including local as well as
global interpretations.

Among all explanation methods, we dwell on the methods which approximate a
black-box model by the linear model. The first well-known method is LIME
(Local Interpretable Model-Agnostic Explanation) \cite{Ribeiro-etal-2016}. It
is based on building a linear model around the explained example. Coefficients
of the linear model are interpreted as the feature's importance. The linear
regression for solving the regression problem or logistic regression for
solving the classification problem allow us to construct the corresponding
linear models by generating many synthetic examples in the vicinity of the
explained example. In fact, LIME can be viewed as a linear approximation of
some non-linear function at a point.

It should be noted that the simple linear model due to its simplicity may
provide inadequate approximation of a black-box model. To overcome this
difficulty a series of models based on applying Generalized Additive Models
(GAMs) \cite{Hastie-Tibshirani-1990} has been developed. GAMs are models which
are represented as a linear combination of some shape functions of features.
The idea of using functions of features instead of the features themselves has
extended the set of possible explanation models. In particular, one of the
explanation methods using GAMs is Explainable Boosting Machine (EBM) proposed
by Nori et al. \cite{Nori-etal-19}. EBM by means of the gradient boosting
machine produces special shape functions which clearly illustrate how features
contribute to the model predictions. However, the most interesting method
based on GAM is the Neural Additive Model (NAM) \cite{Agarwal-etal-20}. It is
implemented as a linear combination of neural networks such that a single
feature is fed to each network, and the output of each network is a shape
function of a feature. In other words, we do not need to invent the function,
it is inferred as a result of the neural network training. The impact of every
feature on the prediction is determined by its corresponding shape function.
Similar methods using neural networks for constructing GAMs and performing
shape functions called GAMI-Net and the Adaptive Explainable Neural Networks
(AxNNs) were proposed by Yang et al. \cite{Yang-Zhang-Sudjianto-20} and Chen
et al. \cite{Chen-Vaughan-etal-20}, respectively.

LIME as well as other methods have been successfully applied to many machine
learning models for explanation. However, to the best of our knowledge, there
is a large class of models for which there are a few explanation methods.
These models solve the problems in the framework of survival analysis which
can be regarded as a fundamental tool for solving many tasks characterizing by
the \textquotedblleft time to event\textquotedblright\ data
\cite{Hosmer-Lemeshow-May-2008}. There are many machine learning survival
models trained on datasets consisting of censored and uncensored data and
predicting probabilistic measures in accordance with the feature values of an
object analyzed \cite{Wang-Li-Reddy-2017}. One of the popular semi-parametric
regression models for analysis of survival data, which takes into account
features of training examples, is the Cox proportional hazards model, which
calculates effects of observed covariates (features) on the risk of an event
occurring, for example, the death or failure \cite{Cox-1972}. It should be
noted that the Cox model plays a crucial role in the explanation models
because it considers a linear combination of the example covariates.
Therefore, coefficients of the covariates can be regarded as quantitative
impacts on the prediction. In fact, the Cox model is used as the linear
approximation of the machine learning models. In spite of importance of
survival models, especially in medicine, we have to admit that there are only
extensions of LIME to explain the survival model predictions
\cite{Kovalev-Utkin-2020c,Kovalev-Utkin-Kasimov-20a,Utkin-Kovalev-Kasimov-20c}%
. These methods called SurvLIME are based on the linear Cox model to produce
explanations. However, it is interesting to extend bounds which restrict the
use of the Cox model as the model with linear combination of covariates. This
extension can be implemented by using GAM instead of the linear relationship
in the Cox model and by modifying NAM to train the neural network aiming to
approximate the extended Cox model and the survival machine learning model
(the black box).

Therefore, a new method called Survival NAM (SurvNAM) is proposed as a
combination of the survival model explanation methods
\cite{Kovalev-Utkin-2020c,Kovalev-Utkin-Kasimov-20a,Utkin-Kovalev-Kasimov-20c}
and the NAM method \cite{Agarwal-etal-20}. The basic idea behind the
combination is to construct an expected loss function for training the NAM
taking into account peculiarities of survival analysis and survival machine
learning models. We also propose two interesting modifications of SurvNAM. The
first modification is based on using the Lasso-based regularization for shape
functions from GAM. Its use produces a sparse representation of GAM and
prevents the neural network from overfitting which may take place due to
influence of unimportant features on the training process of the neural
network. Another modification is based on the idea to add a linear component
to each shape function $g_{k}$ in GAM with some weights which are training
parameters. This idea relates to the shortcut connection in the well-known
ResNet \cite{He-Zhang-Ren-Sun-2016}, which plays an important role in
alleviating the vanishing gradient problem \cite{Liu-Chen-etal-2019}.

The paper is organized as follows. Related work can be found in Section 2. A
short description of NAM as well as basic concepts of survival analysis,
including the Cox model, is given in Section 3. Basic ideas behind the
proposed SurvNAM method are briefly considered in Section 4. A formal
derivation of the convex expected loss function for training the neural
network in NAM can be found in Section 5. Modifications of SurvNAM are
presented in Section 6. Numerical experiments with synthetic and real data are
provided in Section 7. A discussion comparing SurvNAM and SurvLIME
\cite{Kovalev-Utkin-2020c,Kovalev-Utkin-Kasimov-20a,Utkin-Kovalev-Kasimov-20c}
can be found in Section 8. Concluding remarks are provided in Section 9.

\section{Related work}

\textbf{Local and global explanation methods.} Many methods have been
developed to locally explain black-box models. The first one is the original
LIME method \cite{Ribeiro-etal-2016}. Its success and simplicity motivated to
develop several modifications, for example, ALIME
\cite{Shankaranarayana-Runje-2019}, NormLIME \cite{Ahern-etal-2019}, DLIME
\cite{Zafar-Khan-2019}, Anchor LIME \cite{Ribeiro-etal-2018}, LIME-SUP
\cite{Hu-Chen-Nair-Sudjianto-2018}, LIME-Aleph \cite{Rabold-etal-2019},
GraphLIME \cite{Huang-Yamada-etal-2020}. Modifications of LIME to explain the
survival model predictions were proposed in
\cite{Kovalev-Utkin-2020c,Kovalev-Utkin-Kasimov-20a,Utkin-Kovalev-Kasimov-20c}.

Another powerful method is the SHAP \cite{Strumbel-Kononenko-2010} which takes
a game-theoretic approach for optimizing a regression loss function based on
Shapley values \cite{Lundberg-Lee-2017}. In contrast to LIME which only
locally explains, the SHAP method can be used for global explanation. There
are also alternative methods which include influence functions
\cite{Koh-Liang-2017}, a multiple hypothesis testing framework
\cite{Burns-etal-2019}, and many other methods.

A set of quite different explanation methods can be united as counterfactual
explanations \cite{Wachter-etal-2017}. According to \cite{Molnar-2019}, a
counterfactual explanation of a prediction can be defined as the smallest
change to the feature values of an input original example that changes the
prediction to a predefined outcome. Due to intuitive and human-friendly
explanations provided by this family of methods, several its modifications has
been proposed
\cite{Goyal-etal-2018,Hendricks-etal-2018,Looveren-Klaise-2019,Waa-etal-2018}.
Counterfactual modifications of LIME have been also proposed by Ramon et al.
\cite{Ramon-etal-2020} and White and Garcez \cite{White-Garcez-2020}.

Many aforementioned explanation methods are based on perturbation techniques
\cite{Fong-Vedaldi-2019,Fong-Vedaldi-2017,Petsiuk-etal-2018,Vu-etal-2019}.
These methods assume that contribution of a feature can be determined by
measuring how prediction score changes when the feature is altered
\cite{Du-Liu-Hu-2019}. One of the advantages of perturbation techniques is
that they can be applied to a black-box model without any need to access the
internal structure of the model. A possible disadvantage of perturbation
technique is its computational complexity when perturbed input examples are of
the high dimensionality. Moreover, this technique may meet some difficulties
when categorical features are perturbed.

Descriptions of many explanation methods and various approaches, their
critical reviews can be found in survey papers
\cite{Adadi-Berrada-2018,Arrieta-etal-2019,Carvalho-etal-2019,Guidotti-2019,Rudin-2019,Rudin-etal-21}%
. This is a small part of a large number of reviews devoted to explanation methods.

\textbf{Machine learning models in survival analysis}. An excellent review of
machine learning methods dealing with data in the framework of survival
analysis is presented by Wang et al. \cite{Wang-Li-Reddy-2017}. One of the
popular and simple survival models is the Cox model. Its nice properties
motivated developing many approaches which extend it. Tibshirani
\cite{Tibshirani-1997} proposed a modification of the Cox model based on the
Lasso method in order to take into account a high dimensionality of survival
data. Following this paper, several modifications of the Lasso methods for
censored data were proposed
\cite{Kaneko-etal-2015,Kim-etal-2012,Ternes-etal-2016,Witten-Tibshirani-2010,Zhang-Lu-2007}%
. Many models have been developed to relax the linear relationship assumption
accepted in the Cox model
\cite{Van_Belle-etal-2011,Faraggi-Simon-1995,Haarburger-etal-2018,Katzman-etal-2018,Ranganath-etal-2016,Zhu-Yao-Huang-2016}%
. We would like to pay attention to random survival forests (RSFs) which
became one of the most powerful and efficient tools for the survival analysis
\cite{Ibrahim-etal-2008,Mogensen-etal-2012,Nasejje-etal-2017,Omurlu-etal-2009,Schmid-etal-2016,Wang-Zhou-2017,Wright-etal-2016,Wright-etal-2017}%
.

It should be noted that most models dealing with survival data can be regarded
as black-box models and should be explained. Only the Cox model has a simple
explanation due to its linear relationship between covariates. This property
of the Cox model allows us using this model or its modifications for
explaining complex black-box survival models.

\textbf{GAMs and NAMs in explanation problems}. Since GAM is a more general
and flexible model in comparison with the original linear model, then many
authors used GAM for developing new explanation models approximating the
black-box model for local and global explanations. Several explanation models
using the gradient boosting machines \cite{Friedman-2001,Friedman-2002} to
produce GAMs were proposed by \cite{Lou-etal-12,Zhang-Tan-Koch-etal-19}. An
idea behind their using as interpretation models is that all features are
sequentially considered in each iteration of boosting to learn shape function
of the features. GAMs are a class of linear models where the outcome is a
linear combination of some functions of features. They aim to provide a better
flexibility for the approximation of the black-box model and to determine the
feature importance by analyzing how the feature affects the predicted output
through its corresponding function \cite{Lou-etal-12,Zhang-Tan-Koch-etal-19}.
One of the interpretation methods using GAMs based on the gradient boosting
machines is Explainable Boosting Machine (EBM) proposed by Nori et al.
\cite{Nori-etal-19} and Chang et al. \cite{Chang-Tan-etal-2020}. According to
EBM, shape functions are gradient-boosted ensembles of bagged trees, each tree
operating on a single variable. Another interesting class of models called
NAMs was proposed by Agarwal et al. \cite{Agarwal-etal-20}. NAMs learn a
linear combination of neural networks that trained jointly. NAM can be viewed
as a neural network implementation of GAM where shape functions are selected
from a class of functions which can be realized by neural networks with a
certain architecture. Similar approaches using neural networks for
constructing GAMs and performing shape functions called GAMI-Net and the
Adaptive Explainable Neural Networks (AxNNs) were proposed by Yang et al.
\cite{Yang-Zhang-Sudjianto-20} and Chen et al. \cite{Chen-Vaughan-etal-20},
respectively. In order to avoid the neural network overfitting, an ensemble of
gradient boosting machines producing shape functions was proposed by
Konstantinov and Utkin \cite{Konstantinov-Utkin-21}. In contrast to NAM, the
ensemble provides weights of features in the explicit form, and it is simply trained.

\section{Background}

\subsection{A short description of NAM}

\begin{figure}
[ptb]
\begin{center}
\includegraphics[
height=2.437in,
width=2.9499in
]%
{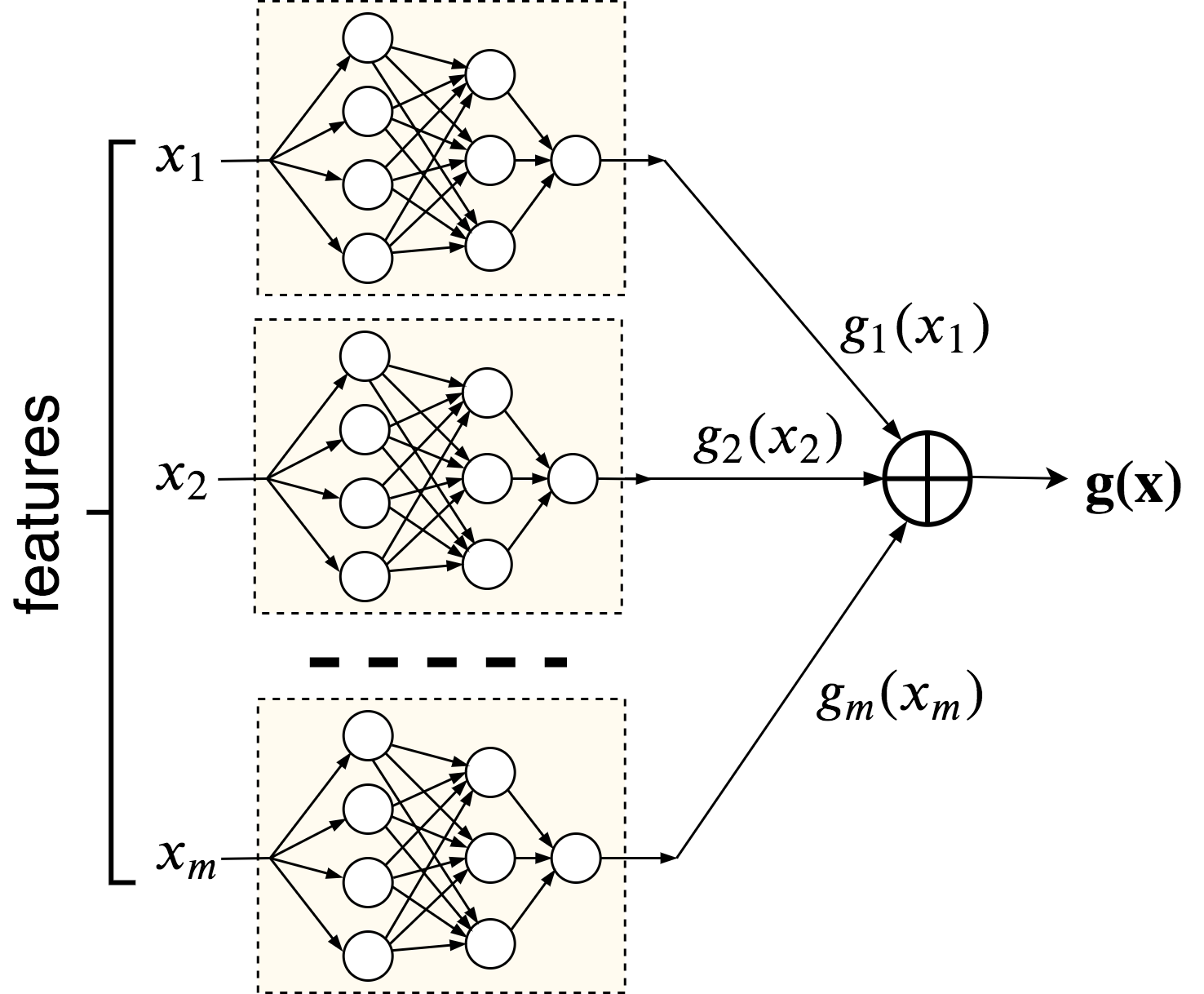}%
\caption{An illustration of NAM}%
\label{fig:NAM_structure}%
\end{center}
\end{figure}

Let us briefly consider NAM proposed by Agarwal et al. \cite{Agarwal-etal-20}.
This is a neural network of a special form shown in Fig.
\ref{fig:NAM_structure}. First of all, an idea of its architecture is based on
applying GAM \cite{Hastie-Tibshirani-1990}, which can be written as follows:
\begin{equation}
g(\mathbb{E}(y(\mathbf{x}))=g_{1}(x_{1})+...+g_{m}(x_{m}),
\label{Interpr_GBM_1}%
\end{equation}
where $\mathbf{x}=(x_{1},...,x_{m})$ is the feature vector; $y$ is the target
variable; $g_{i}$ is a univariate shape function with $\mathbb{E}(g_{i})=0$,
$g$ is a link function (the identity or log functions) relating the expected
value of $y$ to the features.%

GAM allows us to easily establish a partial relationships between the
predicted output $y$ and the features. Moreover, the marginal impact of a
single feature does not depend on the others in the model due to its additive
nature. GAMs provide a better flexibility for the approximation of the
black-box model and determine the feature importance by analyzing how the
feature affects the predicted output through its corresponding function
\cite{Lou-etal-12,Zhang-Tan-Koch-etal-19}.

It can be seen from Fig. \ref{fig:NAM_structure} that NAM consists of $m$
subnetworks such that a single feature is fed to each subnetwork. Outcomes of
subnetworks are values of functions $g_{i}(x_{i})$, $i=1,...,m$, which can be
regarded as impacts of the corresponding features. Denote a vector of the
shape functions as $\mathbf{g}(\mathbf{x})=(g_{m}(x_{m}),...,g_{m}(x_{m}))$.
It should be noted that subnetworks may have different structures, but all
subnetworks are trained jointly using backpropagation and can learn
arbitrarily complex shape functions \cite{Agarwal-etal-20}.

Suppose that the training set $D=\{(\mathbf{x}_{1},y_{1}),...,(\mathbf{x}%
_{n},y_{n})\}$ consists of $n$ examples, where $\mathbf{x}_{i}$ belongs to a
set $\mathcal{X}\subset\mathbb{R}^{m}$ and represents a feature vector
involving $m$ features, and $y_{i}\in\mathbb{R}$ represents the observed
target value for regression or the class label for classification. The
expected loss function $L(\mathbf{W},D)$ of the whole network can be written
as follows:
\begin{align}
L(\mathbf{W},D)  &  =\sum_{i=1}^{n}l\left(  y_{i},\sum_{k=1}^{m}g_{k}%
(x_{k}^{(i)})\right) \nonumber\\
&  =\sum_{i=1}^{n}\left(  y_{i}-\sum_{k=1}^{m}g_{k}(x_{k}^{(i)})\right)
^{2}=\sum_{i=1}^{n}\left(  y_{i}-\mathbf{g}(\mathbf{x}_{i})\cdot
\mathbf{1}^{\mathrm{T}}\right)  ^{2}, \label{SurvNAM_14}%
\end{align}
where $x_{k}^{(i)}$ is the $k$-th feature of the $i$-th example; $l$ is the
loss function which, for example, is the distance between the target value
$y_{i}$ and the network output; $\mathbf{1}$ is the unit vector; $\mathbf{W}$
is a set of the network training parameters.

Details of the whole neural network, including its training, the used
regularization techniques, activation functions, are considered in
\cite{Agarwal-etal-20}.

\subsection{Elements of survival analysis}

The training set in survival analysis consists of censored and uncensored
observations. For censored observations, it is only known that the time to the
event exceeds the duration of observation. The case when the time to the event
coincides with the duration of the observation corresponds to the uncensored
observation. Taking into account these types of observations, we can write the
training set $D$ consisting of $n$ triplets $(\mathbf{x}_{i},\delta_{i}%
,T_{i})$, $i=1,...,n$, such that $\mathbf{x}_{i}^{\mathrm{T}}=(x_{1}%
^{(i)},...,x_{m}^{(i)})$ is the feature vector of the $i$-th example; $T_{i}$
is the time to the event of interest; $\delta_{i}\in\{0,1\}$ is the indicator
of censored or uncensored observations. In particular, $\delta_{i}=1$ when the
event of interest is observed (the uncensored observation). If $\delta_{i}=0$,
then we have the censored observation.

A survival machine learning model is trained on the set $D$ in order to
estimate probabilistic characteristics of time $T$ to the event of interest
for a new instance $\mathbf{x}$.

One of the important concepts in survival analysis is the survival function
(SF) $S(t|\mathbf{x})$ which is the probability of surviving of instance
$\mathbf{x}$ up to time $t$, i.e., $S(t|\mathbf{x})=\Pr\{T>t|\mathbf{x}\}$.
Another important concept is the cumulative hazard function (CHF)
$H(t|\mathbf{x})$, which is interpreted as the probability of an event at time
$t$ given survival until time $t$. The SF $S(t|\mathbf{x})$ is determined
through the CHF $H(t|\mathbf{x})$ as follows:%
\begin{equation}
S(t|\mathbf{x})=\exp\left(  -H(t|\mathbf{x})\right)  .
\end{equation}

The CHF can also be defined as the integral of the hazard function
$h(t|\mathbf{x})$ which is also used in survival analysis and defined as the
rate of event at time $t$ given that no event occurred before time $t$.

An important measure characterizing a survival machine learning model, is the
Harrell's concordance index or the C-index \cite{Harrell-etal-1982}. It can be
used for comparing different models and for tuning their parameters. It can be
also regarded as the prediction error to evaluate the overall fitting. The
C-index measures the probability that, in a randomly selected pair of
examples, the example that fails first had a worst predicted outcome. It is
calculated as the ratio of the number of pairs correctly ordered by the model
to the total number of admissible pairs. A pair is not admissible if the
events are both right-censored or if the earliest time in the pair is
censored. If the C-index is equal to 1, then the corresponding survival model
is supposed to be perfect. If the C-index is 0.5, then the model is not better
than random guessing.

A popular regression model for the analysis of survival data is the well-known
Cox proportional hazards model that calculates the effects of observed
covariates on the risk of an event occurring \cite{Cox-1972}. The model
assumes that the log-risk of an event of interest is a linear combination of
covariates or features. This assumption is referred to as the linear
proportional hazards condition. The Cox model is semi-parametric in the sense
that it can be factored into a parametric part, which consists of a regression
parameter vector associated with the covariates, and a non-parametric part,
which can be left completely unspecified \cite{Devarajn-Ebrahimi-2011}.

In accordance with the Cox model \cite{Cox-1972,Hosmer-Lemeshow-May-2008}, the
hazard function at time $t$ given predictor values $\mathbf{x}$ is defined as
\begin{equation}
h(t|\mathbf{x},\mathbf{b})=h_{0}(t)\exp\left(  \psi(\mathbf{x},\mathbf{b}%
)\right)  . \label{SurvLIME1_10}%
\end{equation}

Here $h_{0}(t)$ is a baseline hazard function which does not depend on the
vector $\mathbf{x}$ and the vector $\mathbf{b}$; $\mathbf{b}^{\mathrm{T}%
}=(b_{1},...,b_{m})$ is an unknown vector of regression coefficients or
parameters. The baseline hazard function represents the hazard when all of the
covariates are equal to zero, i.e., it describes the hazard for zero feature
vector $\mathbf{x}^{\mathrm{T}}=(0,...,0)$. The function $\psi(\mathbf{x}%
,\mathbf{b})$ in the model is linear, i.e.,
\begin{equation}
\psi(\mathbf{x},\mathbf{b})=\mathbf{b}^{\mathrm{T}}\mathbf{x}=\sum
\nolimits_{k=1}^{m}b_{k}x_{k}.
\end{equation}

A hazard ratio is the ratio between two hazard functions $h(t|\mathbf{x}%
_{1},\mathbf{b})$ and $h(t|\mathbf{x}_{2},\mathbf{b})$. It is constant for the
Cox model over time. In the framework of the Cox model, SF $S(t|\mathbf{x}%
,\mathbf{b})$ is computed as
\begin{equation}
S(t|\mathbf{x},\mathbf{b})=\exp(-H_{0}(t)\exp\left(  \psi(\mathbf{x}%
,\mathbf{b})\right)  )=\left(  S_{0}(t)\right)  ^{\exp\left(  \psi
(\mathbf{x},\mathbf{b})\right)  }.
\end{equation}

Here $H_{0}(t)$ is the baseline CHF; $S_{0}(t)$ is the baseline SF which is
defined as the SF for zero feature vector $\mathbf{x}^{\mathrm{T}}=(0,...,0)$.
It is important to note that functions $H_{0}(t)$ and $S_{0}(t)$ do not depend
on $\mathbf{x}$ and $\mathbf{b}$.

One of the main problems of using the Cox model is linear relationship
assumption between covariates and the log-risk of an event. Various
modifications have been proposed to generalize the Cox model taking into
account the corresponding non-linear relationship. The first group of models
uses a neural network for modelling the non-linear function. Following the
pioneering work of Faraggi and Simon \cite{Faraggi-Simon-1995}, where a simple
feed-forward neural network was proposed to model the non-linear relationship
between covariates and the log-risk, many machine learning models have been
proposed to relax this condition, including deep neural networks, the support
vector machine, the random survival forest, etc.
\cite{Haarburger-etal-2018,Katzman-etal-2018,Nezhad-etal-2018,Polsterl-etal-2016,Van_Belle-etal-2011,Wang-Li-Reddy-2017}%
.

One of the ways to extend the Cox model is to replace the function
$\psi(\mathbf{x},\mathbf{b})$ with GAM, i.e., with the function
(\ref{Interpr_GBM_1}). As a result, we get an extension of the Cox model with
the function
\begin{equation}
\psi(\mathbf{x},\mathbf{g})=g_{1}(x_{1})+...+g_{m}(x_{m}).
\end{equation}
The combination of GAM and the Cox model was studied by several authors
\cite{Bender-etal-18,Hastie-Tibshirani-1990,Hastie-Tibshirani-95}.

\section{An overview of the proposed main algorithm}

If the have a training set $D$ and a black-box survival machine learning model
trained on $D$, then the black-box model with input $\mathbf{x}$ produces the
corresponding output in the form of the CHF $H(t|\mathbf{x})$ or the SF
$S(t|\mathbf{x})$. The basic idea behind the explanation algorithm SurvNAM is
to approximate the black-box model by the extension of the Cox model with GAM
such that GAM is implemented by means of NAM trained by applying a specific
loss function taking into account peculiarities of the Cox model and the
black-box survival model output which is represented in the form of the SF or
the CHF. In other words, we try to learn NAM for getting univariate shape
functions $g_{i}$ whose sum is the function $\psi(\mathbf{x},\mathbf{g})$. For
implementing the above idea, we have to define an approximation quality
measure of the black-box model and the extension of the Cox model outputs and
then to apply this measure to constructing the expected loss function
$L(\mathbf{W},D)$ for training NAM.

Let us denote the black-box survival model output corresponding to the input
vector $\mathbf{x}$ as $H(t|\mathbf{x})$ and the output of the Cox model
extension as $H^{\ast}(t|\mathbf{x},\mathbf{g})$. Then the approximation
quality measure mentioned above can be defined as the distance $d$ between
$H(t|\mathbf{x})$ and $H^{\ast}(t|\mathbf{x},\mathbf{g})$.

Let us determine the training procedure of NAM depending on a type of
explanation. If we aim to get the \textit{global explanation}, then the whole
dataset $D$ can be used. However, if we have to implement the \textit{local
explanation} for a point $\mathbf{x}$, then $N$ points $\mathbf{x}%
_{1},...,\mathbf{x}_{N}$ are randomly generated around $\mathbf{x}$ in
accordance with a predefined probability distribution function. These examples
are fed to the black-box survival model which produces a set of the
corresponding CHFs $H(t|\mathbf{x}_{1}),...,H(t|\mathbf{x}_{N})$. In sum, the
training set for learning NAM in the case of the local explanation is the set
of the generated points and the corresponding CHFs.

A criterion of the approximation quality is the weighted mean of the distances
$d$ between CHFs over all points $\mathbf{x}_{1},...,\mathbf{x}_{N}$ where
each weight depends on the distance between generated point $\mathbf{x}_{k}$
and explained point $\mathbf{x}$. Smaller distances between $\mathbf{x}_{k}$
and $\mathbf{x}$ produce larger weights of distances between CHFs.

In fact, we have a double approximation. First, we implicitly approximate the
black-box survival model by the extended Cox model. This approximation is
implicit because its result is the loss function taking into account the
distance between the black-box survival model prediction and the extended Cox
model prediction. Second, we learn the neural network (NAM) to find shape
functions approximating the extended Cox model.

\begin{figure}
[ptb]
\begin{center}
\includegraphics[
height=4.7729in,
width=4.1554in
]%
{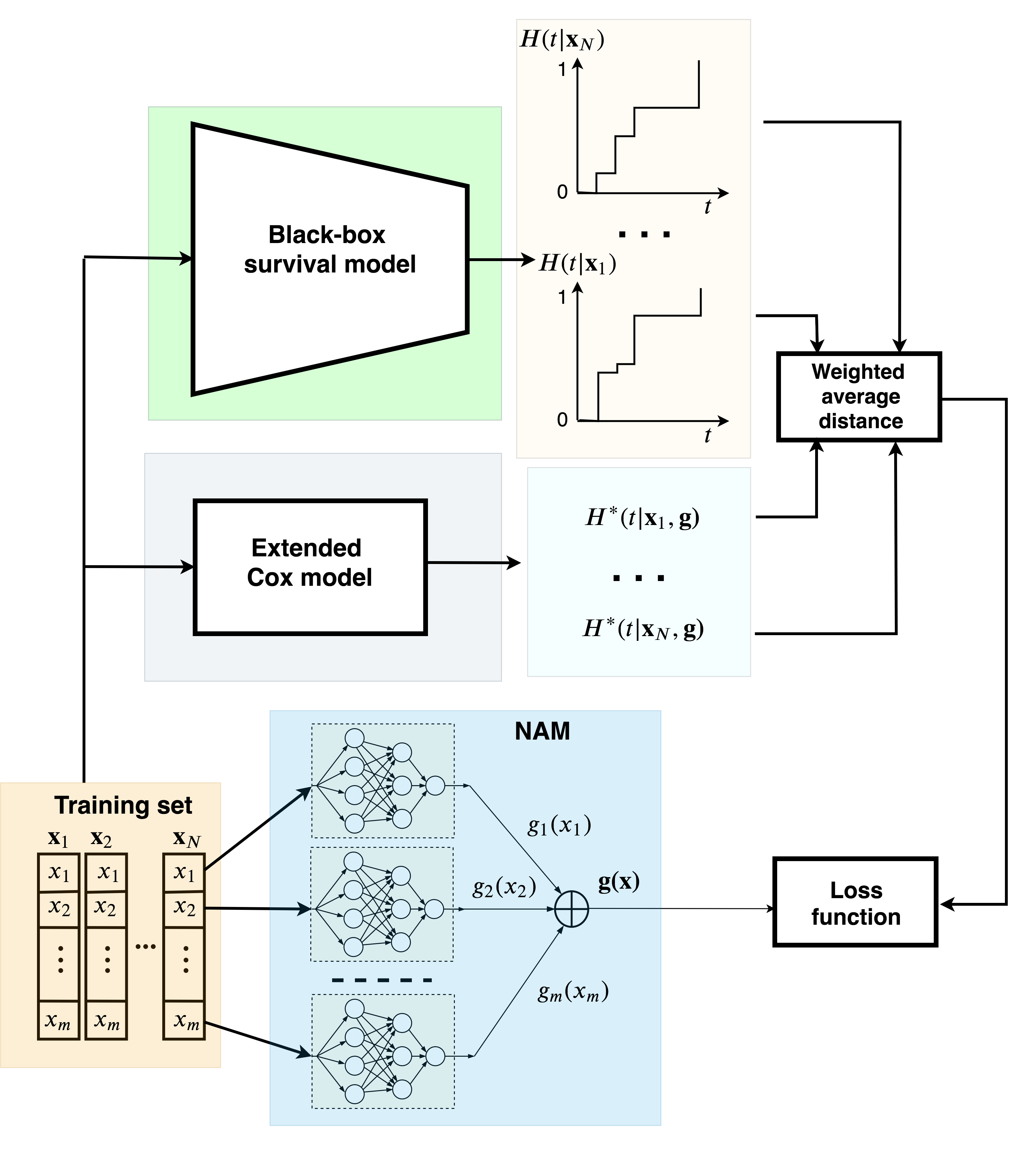}%
\caption{A schematic illustration of SurvNAM}%
\label{f:survnam_explain}%
\end{center}
\end{figure}

A general scheme of SurvNAM is shown in Fig. \ref{f:survnam_explain}. The
black-box survival model is trained on $N$ vectors of training examples such
that each example consists of $m$ features. Predictions of the model are CHFs.
In accordance with the Cox model, the baseline CHF is estimated, and the
predictions of the extended Cox models are represented as functions of the
training vectors $\mathbf{x}_{i}$ and the vector of shape functions
$\mathbf{g}(\mathbf{x}_{i})$. The weighted average distance between
predictions of the black-box survival model and the prediction representation
of the extended Cox model compose the loss function for training NAM. By
having the trained NAM, we get shape functions $g_{1}(x_{1}),...,g_{m}(x_{m})$
for all features. These functions can be viewed as explanation of the
predictions and the corresponding survival model.%

\section{The expected loss function}

The loss function for training NAM differs from the standard loss function
like (\ref{SurvNAM_14}). First, it has to take into account the peculiarity of
the black-box survival model output which is a function (the CHF or the SF).
Second, it has to take into account peculiarities of the Cox model extension.
Third, the loss function should be convex to get optimal solution.

Before deriving the loss function, we introduce some notations and conditions.

Let $t_{0}<t_{1}<...<t_{s}$ be distinct times to events of interest, for
example, times to the patient deaths from the set $\{T_{1},...,T_{n}\}$, where
$t_{0}=\min_{k=1,...,n}T_{k}$ and $t_{s}=\max_{k=1,...,n}T_{k}$. The black-box
model maps the feature vectors $\mathbf{x}\in\mathbb{R}^{m}$ into piecewise
constant CHFs $H(t|\mathbf{x})$. It is assumed that $\max_{t}H(t|\mathbf{x}%
)<\infty$.

Let us introduce the time $T=t_{s}+\gamma$ in order to restrict the integral
of $H(t|\mathbf{x})$, where $\gamma$ is a small positive number. Let
$\Omega=[0,T]$. Then we can write $\int_{\Omega}H(t|\mathbf{x})\mathrm{d}%
t<\infty$. Since the CHF $H(t|\mathbf{x})$ is piecewise constant, then it can
be written in a special form. Let us divide the set $\Omega$ into $s+1$
subsets $\Omega_{0},...,\Omega_{s}$ such that $\Omega=\cup_{j=0,...,s}%
\Omega_{j}$, where $\Omega_{s}=[t_{s},T]$, $\Omega_{j}=[t_{j},t_{j+1})$,
$\forall j\in\{0,...,s-1\}$, and $\Omega_{j}\cap\Omega_{k}=\emptyset$ for
$\forall j\neq k$

The CHF $H(t|\mathbf{x})$ can be expressed through the indicator functions as
follows:
\begin{equation}
H(t|\mathbf{x})=\sum_{j=0}^{s}H_{j}(\mathbf{x})\cdot\chi_{\Omega_{j}}(t),
\label{SurvNAM_25}%
\end{equation}
where $\chi_{\Omega_{j}}(t)$ is the indicator function taking value $1$ if
$t\in\Omega_{j}$, and $0$ otherwise; $H_{j}(\mathbf{x})$ is a part of the CHF
in interval $\Omega_{j}$, which is constant in $\Omega_{j}$ and does not
depend on $t$.

Let us write similar expressions for the CHF of the Cox model extension:
\begin{equation}
H^{\ast}(t|\mathbf{x},\mathbf{g})=H_{0}(t)\exp\left(  \psi(\mathbf{x}%
,\mathbf{g})\right)  =\sum_{j=0}^{s}\left[  H_{0j}\exp\left(  \psi
(\mathbf{x},\mathbf{g})\right)  \right]  \chi_{\Omega_{j}}(t).
\label{SurvNAM_27}%
\end{equation}

By having the above expressions for the $H(t|\mathbf{x})$ and $H^{\ast
}(t|\mathbf{x},\mathbf{g})$, we can construct the loss function as the
weighted Euclidean distance between $H(t|\mathbf{x}_{i})$ and $H^{\ast
}(t|\mathbf{x}_{i},\mathbf{g}_{i})$ for all generated $\mathbf{x}_{i}$,
$i=1,...,N$:
\[
L(\mathbf{W},D)=\sum_{i=1}^{N}v_{i}\int_{\Omega}\left\Vert H(t|\mathbf{x}%
_{i})-H^{\ast}(t|\mathbf{x}_{i},\mathbf{g}_{i})\right\Vert _{2}^{2}%
\mathrm{d}t,
\]
where the integral is taken over the set $\Omega$ because we define the
distance between functions of time $t$; $v_{i}$ is the weight depending on the
distance between generated point $\mathbf{x}_{i}$ and explained point
$\mathbf{x}$; $\mathbf{x}_{i}$ is the generated point.

Since the CHFs are represented as piecewise constant function of the form
(\ref{SurvNAM_25}) and (\ref{SurvNAM_27}), then it is simply to prove that the
integral in the loss function can be replaced by the sum as follows:
\begin{equation}
L(\mathbf{W},D)=\sum_{i=1}^{N}v_{i}\sum_{j=0}^{s}\left(  H_{j}(\mathbf{x}%
_{i})-\left[  H_{0j}\exp\psi(\mathbf{x}_{i},\mathbf{g}_{i})\right]  \right)
^{2}\tau_{j}. \label{SurvNAM_34}%
\end{equation}
where $\tau_{j}=t_{j+1}-t_{j}$.

The main difficulty of using the CHFs is that the loss function $L(\mathbf{W}%
,D)$ may be non-convex. Therefore, we replace CHFs with their logarithms and
find distances between logarithms of CHFs. Since the logarithm is a monotone
function, then there holds:
\begin{equation}
L(\mathbf{W},D)=\sum_{i=1}^{N}v_{i}\sum_{j=0}^{s}\left(  \ln H_{j}%
(\mathbf{x}_{i})-\ln H_{0j}-\psi(\mathbf{x}_{i},\mathbf{g}_{i})\right)
^{2}\tau_{j}. \label{SurvNAM_35}%
\end{equation}

Let us introduce the following notation for short:
\begin{equation}
\Phi_{j}(\mathbf{x}_{i})=\ln H_{j}(\mathbf{x}_{i})-\ln H_{0j}.
\end{equation}

Hence, we get the final expected loss function
\begin{equation}
L(\mathbf{W},D)=\sum_{i=1}^{N}v_{i}\sum_{j=0}^{s}\left(  \Phi_{j}%
(\mathbf{x}_{i})-\sum_{k=1}^{m}g_{k}(x_{k}^{(i)})\right)  ^{2}\tau_{j}.
\label{SurvNAM_36}%
\end{equation}

Parameters $\mathbf{W}$ of the neural network are implicitly contained in
calculating functions $g_{k}$. By adding the regularization term to the loss
function (\ref{SurvNAM_36}), we get
\[
L(\mathbf{W},D)=\sum_{i=1}^{N}v_{i}\sum_{j=0}^{s}\left(  \Phi_{j}%
(\mathbf{x}_{i})-\sum_{k=1}^{m}g_{k}(x_{k}^{(i)})\right)  ^{2}\tau_{j}.
\]

We also assume that $H_{j}(\mathbf{x})\geq\varepsilon>0$ for all $\mathbf{x}$
in order to avoid the case when the logarithm of zero is calculated. It should
be noted that the loss function (\ref{SurvNAM_35}) is not equivalent to the
previous loss function (\ref{SurvNAM_34}). However, this is not so important
because we train the neural network which is fitted to the changed loss
function. Since $\psi(\mathbf{x}_{i},\mathbf{g}_{i})$ is the sum of the
functions $\mathbf{g}_{i}$ which are outputs of NAM, then $\ln H_{j}%
(\mathbf{x}_{i})-\ln H_{0j}-\sum_{k=1}^{m}g_{k}(x_{k}^{(i)})$ in
(\ref{SurvNAM_36}) is the linear function of shape functions $g_{k}%
(x_{k}^{(i)})$. It follows from the above and from positivity of $\tau_{j}$
and $v_{i}$ that the expected loss function (\ref{SurvNAM_36}) is convex. The
weight $v_{i}$ is taken as a decreasing function of the distance between
$\mathbf{x}$ and $\mathbf{x}_{i}$, for example, $v_{k}=K(\mathbf{x}%
,\mathbf{x}_{k})$, where $K(\cdot,\cdot)$ is a kernel.

In sum, we have the expected convex loss function (\ref{SurvNAM_36}) for
training NAM, which can be written as%
\[
L(\mathbf{W},D)=\sum_{i=1}^{N}\sum_{j=0}^{s}\left(  \sqrt{v_{i}\tau_{j}}%
\Phi_{j}(\mathbf{x}_{i})-\sqrt{v_{i}\tau_{j}}\sum_{k=1}^{m}g_{k}(x_{k}%
^{(i)})\right)  ^{2}.
\]

\section{Lasso-based modifications of SurvNAM for the local explanation}

The main difficulty of the proposed approach is that neural subnetworks
computing shape functions $g_{k}$, $k=1,...,m$, may be overfitted due to the
noise from unimportant features. As a result, the approximation may be
unsatisfactory, and important features may be incorrectly ranked. In order to
overcome this difficulty, a sparse modification of SurvNAM is proposed by
using the Lasso method or the $L_{1}$-norm for regularization.

Let us write the GAM as follows:%

\begin{equation}
g(\mathbb{E}(y(\mathbf{x}))=\beta_{1}g_{1}(x_{1})+...+\beta_{m}g_{m}(x_{m}),
\end{equation}
where $\beta_{k}\in\mathbb{R}$ is a coefficient.

It is assumed here that the $k$-th shape function $g_{k}$ has some coefficient
$\beta_{k}$ which determines the impact of function $g_{k}$ on the prediction.
The constraint that $\mathbf{\beta}$ lies in the $L_{1}$-ball encourages
sparsity of the estimated $\mathbf{\beta}$ \cite{Ravikumar-etal-09}. This
implies that if we add the regularization term $\left\Vert \mathbf{\beta
}\right\Vert _{1}=\sum_{k=1}^{m}\left\vert \beta_{k}\right\vert $, then we can
expect that a part of coefficients $\beta_{k}$, corresponding to unimportant
features, will be $0$ or close to $0$ due to properties of the regularization.
Hence, the neural subnetworks, corresponding to unimportant features, will not
introduce the additional noise to results. It should be noted that the sparse
GAM has been studied by several authors
\cite{Avalos-etal-03,Haris-etal-2019,Ravikumar-etal-09}.

The loss function (\ref{SurvNAM_36}) for training the neural network can be
rewritten in accordance with the idea of using the regularization term
$\left\Vert \mathbf{\beta}\right\Vert _{1}$ as follows:%

\begin{align}
L(\mathbf{W},D)  &  =\sum_{i=1}^{N}v_{i}\sum_{j=0}^{s}\left(  \Phi
_{j}(\mathbf{x}_{i})-\sum_{k=1}^{m}\beta_{k}g_{k}(x_{k}^{(i)})\right)
^{2}\tau_{j}\nonumber\\
&  +\lambda\sum_{k=1}^{m}\left\vert \beta_{k}\right\vert , \label{SurvNAM_51}%
\end{align}
where $\lambda$ is a hyper-parameter which controls the strength of the regularization.

In fact, every neural subnetwork is assigned by the weight $\beta_{k}$. It
should be pointed out that the use of the $L_{1}$-norm as the regularization
term in SurvNAM for the sparse explanation differs from use of the original
Lasso method. We use functions $g_{k}$ instead of variables $x_{k}$ here,
which can generally be non-linear. It should be also pointed out that the
visual analysis of shape functions $g_{k}$ is replaced with the analysis of
functions $\beta_{k}g_{k}$, $k=1,...,m$. Numerical experiments with the above
modification of SurvNAM show that the main difficulty of its use is that the
functions $g_{k}(x_{k}^{(i)})$ may increase in order to balance small values
of $\beta_{k}$. A simple way to overcome this difficulty is to restrict the
set of possible functions $g_{k}$. Since function $g_{k}$ is represented as
some function implemented by the $k$-th subnetwork, then its restriction
depends on weights of the subnetwork. Therefore, the set of functions $g_{k}$
can be restricted by regularizing all weights $w_{i}$ of the network
connections. Hence, we add the term $\mu\left\Vert \mathbf{W}\right\Vert
_{2}^{2}$ to the loss function, where $\left\Vert \mathbf{W}\right\Vert
_{2}^{2}$ is the short notation of the squared $L_{2}$-norm for all training
parameters $w_{i}$ of the neural network, $\mu$ is a hyper-parameter which
controls the strength of the regularization.

\begin{figure}
[ptb]
\begin{center}
\includegraphics[
height=1.5056in,
width=3.3754in
]%
{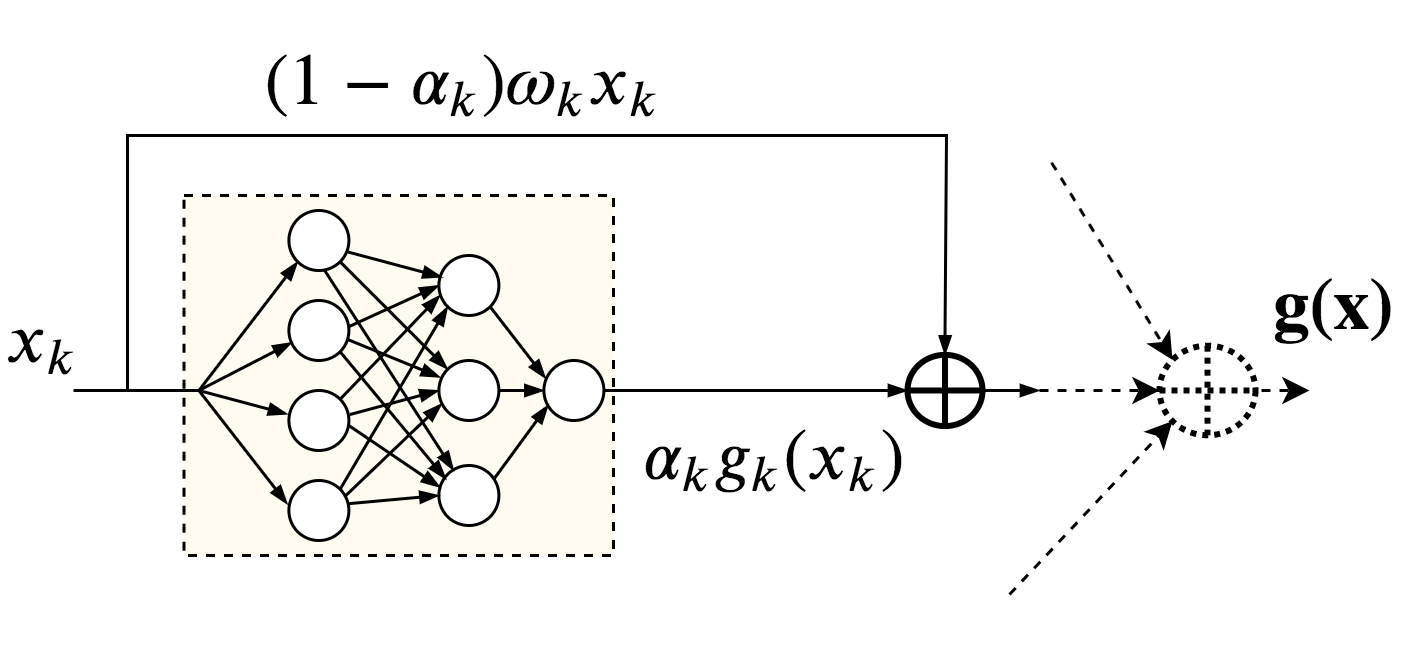}%
\caption{The shortcut connection with regard to the modification of SurvNAM}%
\label{f:shortcut}%
\end{center}
\end{figure}

Another modification is based on the idea to add a linear component to each
function $g_{k}$, i.e., we replace the product $\beta_{k}g_{k}(x_{k}^{(i)})$
with the sum $\alpha_{k}g_{k}(x_{k}^{(i)})+(1-\alpha_{k})\omega_{k}x_{k}%
^{(i)}$. This is an interesting idea because it formally relates to the
shortcut connection in the well-known ResNet proposed by He et al.
\cite{He-Zhang-Ren-Sun-2016}. This connection is skipping one or more layers
in the neural network. It plays an important role in alleviating the vanishing
gradient problem \cite{Liu-Chen-etal-2019}. The shortcut connection with
regard to the $k$-th neural subnetwork in the modification of SurvNAM is
depicted in Fig. \ref{f:shortcut}.%

The aforementioned relationship between the proposed modification of SurvNAM
and the ResNet is very surprising. The difference is that the shortcut
connection in SurvNAM has the weight $(1-\alpha_{k})\omega_{k}$ which consists
of two training parameters $\alpha_{k}$ and $\omega_{k}$. The parameter
$\alpha_{k}$ can be viewed as a measure of the $k$-th feature contribution
non-linearity. The term $(1-\alpha_{k})$ is a measure of the $k$-th feature
contribution linearity. The greater the value of $(1-\alpha_{k})$, the greater
linearity of the $k$-th feature and the smaller the impact of $g_{k}$. An idea
behind the use of parameter $\omega_{k}$ is based on the following reasoning.
Parameter $\alpha_{k}$ chooses between the subnetwork (function $g_{k}$) and
the linear function $x_{k}$. Therefore, $\alpha_{k}$ cannot play a role of an
angle of inclination of the line $x_{k}$. This implies that we add parameter
$\omega_{k}$ to take into account this angle. In other words, we can represent
the sum $\alpha_{k}g_{k}(x_{k})+(1-\alpha_{k})\omega_{k}x_{k}$ as $\alpha
_{k}g_{k}(x_{k})+(1-\alpha_{k})r_{k}$, where $r_{k}=\omega_{k}x_{k}$.

Taking into account restrictions for the network parameters in the form of the
$L_{2}$ regularization $\left\Vert \mathbf{W}\right\Vert _{2}^{2}$, we write
the loss function for training the whole neural network as follows:
\begin{align}
L(\mathbf{W},D)  &  =\sum_{i=1}^{N}v_{i}\sum_{j=0}^{s}\left(  \Phi
_{j}(\mathbf{x}_{i})-\sum_{k=1}^{m}\alpha_{k}g_{k}(x_{k}^{(i)})-\sum_{k=1}%
^{m}(1-\alpha_{k})\omega_{k}x_{k}^{(i)}\right)  ^{2}\tau_{j}\nonumber\\
&  +\lambda\sum_{k=1}^{m}\left\vert \alpha_{k}\right\vert +\mu\cdot\left\Vert
\mathbf{W}\right\Vert _{2}^{2}. \label{SurvNAM_53}%
\end{align}

In contrast to the ResNet, the shortcut connection in SurvNAM plays a double
role. First, it alleviates the vanishing gradient problem when the network is
training. Second, it allows us to see how the linear part of the shape
function impacts on the prediction.

\section{Numerical experiments}

SurvNAM is tested on five real benchmark datasets. A short introduction of the
benchmark datasets are given below.

The \textbf{German Breast Cancer Study Group 2 (GBSG2) Dataset }contains
observations of 686 women. Every example is characterized by 10 features,
including age of the patients in years, menopausal status, tumor size, tumor
grade, number of positive nodes, hormonal therapy, progesterone receptor,
estrogen receptor, recurrence free survival time, censoring indicator (0 -
censored, 1 - event). The dataset can be obtained via the \textquotedblleft
TH.data\textquotedblright\ R package.

The \textbf{Monoclonal Gammopathy (MGUS2) Dataset} contains the natural
history of 1384 subjects with monoclonal gammopathy of undetermined
significance characterized by 5 features. The dataset can be obtained via the
\textquotedblleft survival\textquotedblright\ R package.

The \textbf{Survival from Malignant Melanoma (Melanoma) Dataset} consist of
measurements made on patients with malignant melanoma. The dataset contains
data on 205 patients characterized by 7 features. The dataset can be obtained
via the \textquotedblleft boot\textquotedblright\ R package.

The \textbf{Stanford Heart Transplant (Stanford2) Dataset }contains data on
survival of 185 patients on the waiting list for the Stanford heart transplant
program. The dataset can be obtained via the \textquotedblleft
survival\textquotedblright\ R package.

The \textbf{Veterans' Administration Lung Cancer Study (Veteran) Dataset}
contains data on 137 males with advanced inoperable lung cancer. The subjects
were randomly assigned to either a standard chemotherapy treatment or a test
chemotherapy treatment. Several additional variables were also measured on the
subjects. The dataset can be obtained via the \textquotedblleft
survival\textquotedblright\ R package.

For the local explanation, the perturbation technique is used. In accordance
with the technique, $N$ nearest points $\mathbf{x}_{k}$ are generated in a
local area around the explained example $\mathbf{x}$. These points are
normally distributed with the center at point $\mathbf{x}$ and the standard
deviation which is equal to $10\%$ of the largest distance between points of
the corresponding studied dataset. In numerical experiments, $N=100$. The
weight to the $k$-th point $\mathbf{x}_{k}$ is assigned as follows:
\begin{equation}
w_{k}=1-\sqrt{\frac{\left\Vert \mathbf{x}-\mathbf{x}_{k}\right\Vert _{2}}{r}%
},\ k=1,...,N. \label{SurvLIME_60}%
\end{equation}

As a black-box model, we use the RSF model \cite{Ibrahim-etal-2008} consisting
of $500$ decision survival trees. The approximating Cox model has the baseline
CHF $H_{0}(t)$ constructed on training data using the Nelson--Aalen estimator.

The Python code of NAM is taken from
https://github.com/google-research/google-research/tree/master/neural\_additive\_models.
It is modified to
implement the proposed methods.

\subsection{Numerical experiments with SurvNAM}

\begin{figure}
[ptb]
\begin{center}
\includegraphics[
height=1.9355in,
width=3.6737in
]%
{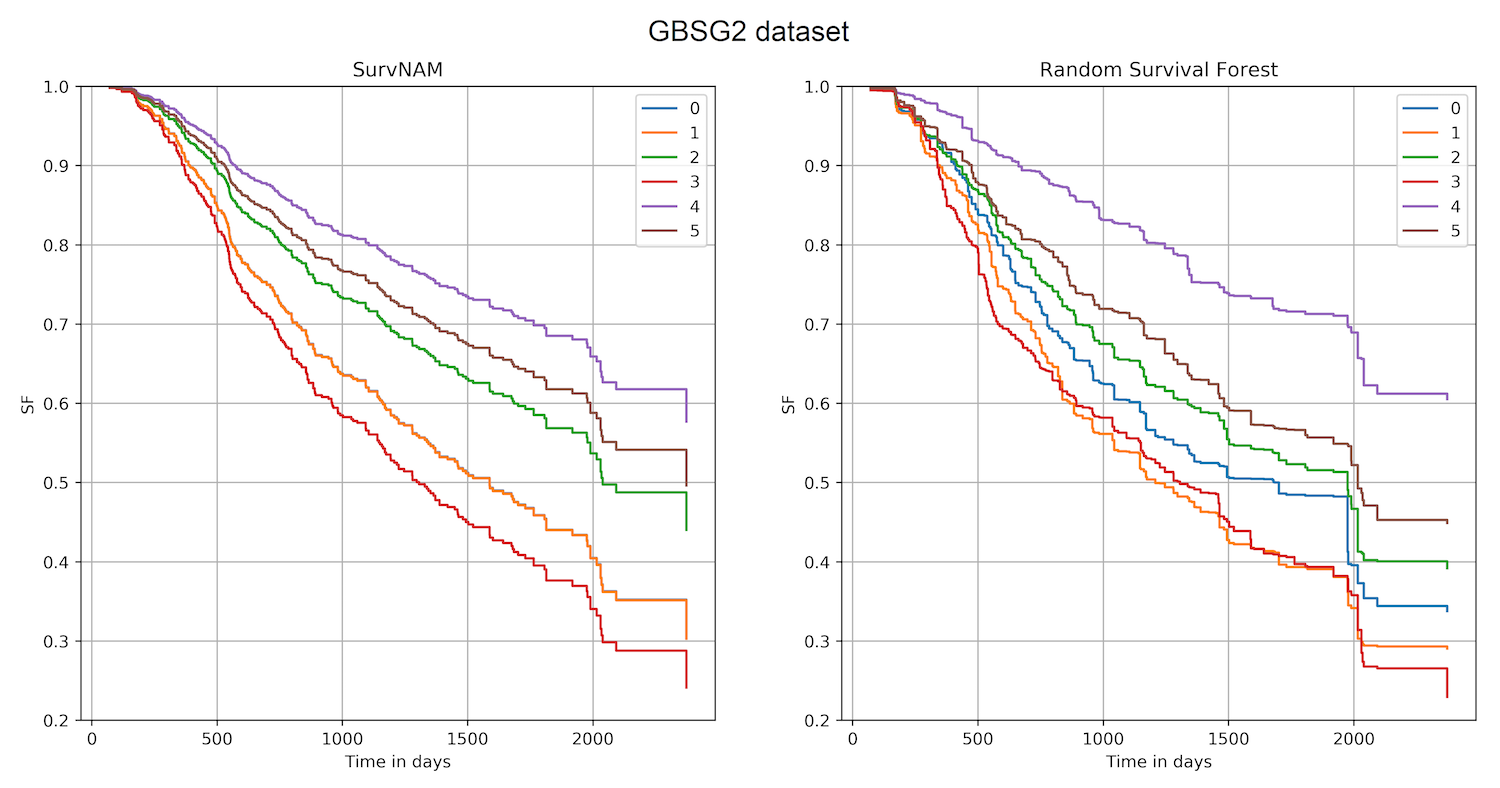}%
\caption{SFs obtained by means of SurvNAM (left picture) and the black-box RSF
model (right picture) for the dataset GBSG2 and the local explanation}%
\label{f:gbsg-2_local_sf}%
\end{center}
\end{figure}

\begin{figure}
[ptb]
\begin{center}
\includegraphics[
height=1.9268in,
width=3.6789in
]%
{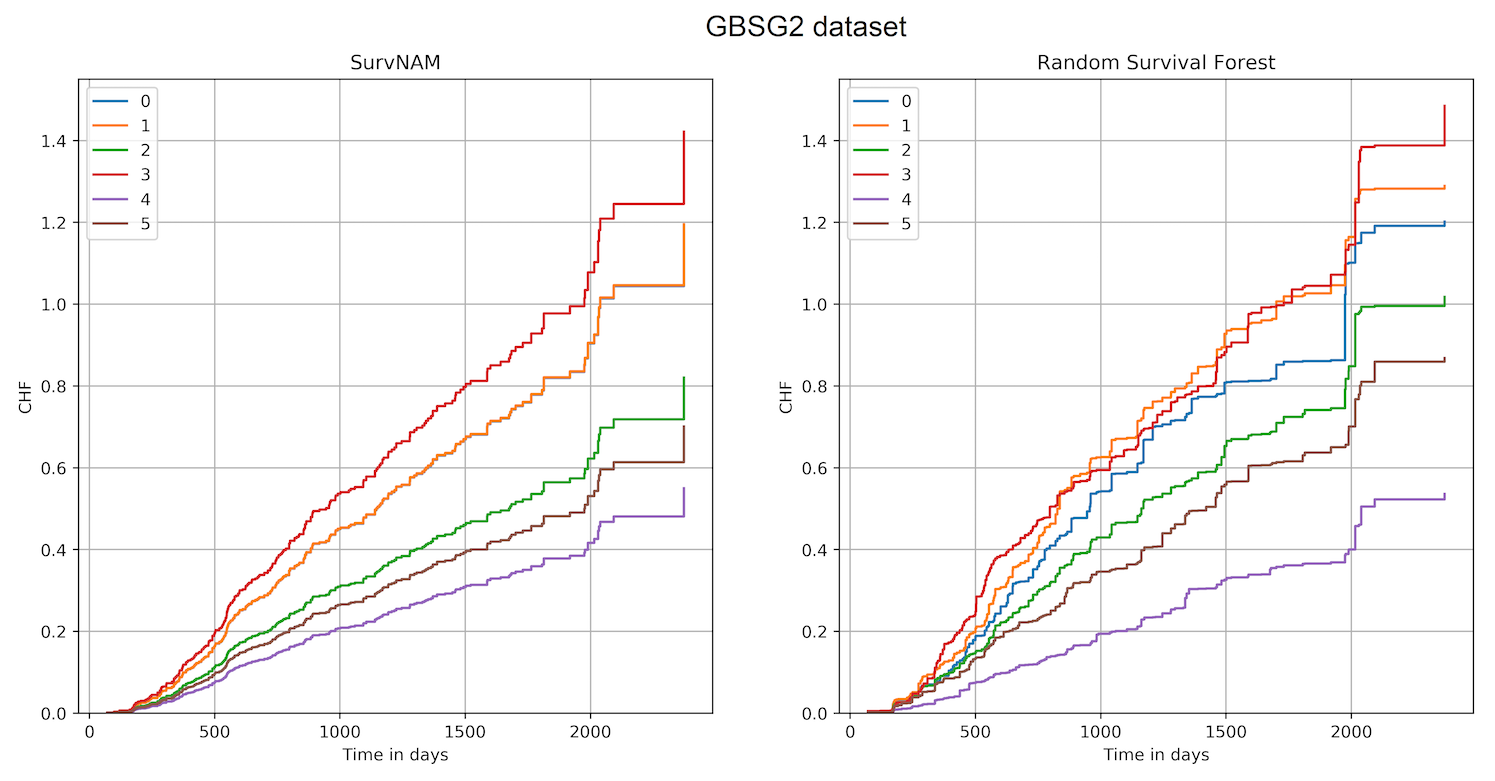}%
\caption{CHFs obtained by means of SurvNAM (left picture) and the black-box
RSF model (right picture) for the dataset GBSG2 and the local explanation}%
\label{f:gbsg-2_local_chf}%
\end{center}
\end{figure}

\begin{figure}
[ptb]
\begin{center}
\includegraphics[
height=2.4414in,
width=3.6556in
]%
{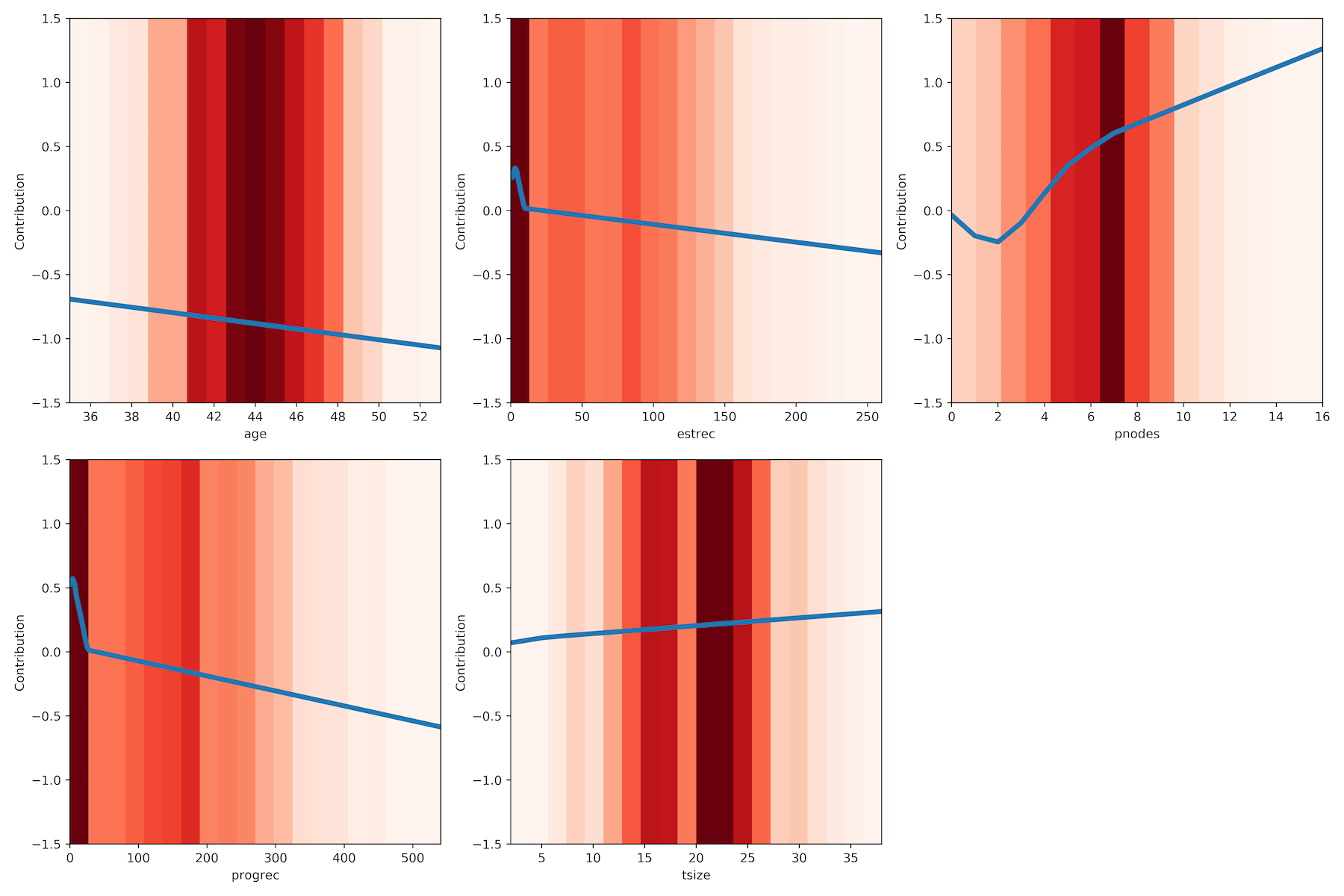}%
\caption{Shape functions for every feature obtained by means of SurvNAM for
the dataset GBSG2 and the local explanation}%
\label{f:gbsg-2_local_shape}%
\end{center}
\end{figure}

First, we test SurvNAM trained on the GBSG2 dataset for the local explanation.
Fig. \ref{f:gbsg-2_local_sf} illustrates SFs obtained by means of the RSF for
$6$ randomly selected examples from the training set (the right picture) and
by means of the extended Cox model with the shape functions $g_{i}(x_{i})$ of
the GAM provided by SurvNAM (the left picture). SFs shown in the left picture
of Fig. \ref{f:gbsg-2_local_sf} can be viewed as approximations approximating
of the corresponding SFs depicted in the right picture. The same CHFs obtained
by the RSF (the right picture) and by the extended Cox model (the left
picture) are shown in Fig. \ref{f:gbsg-2_local_chf}. Shape functions for all
features are illustrated in Fig. \ref{f:gbsg-2_local_shape}. They are computed
for an example with features \textquotedblleft age\textquotedblright=44,
\textquotedblleft estrec\textquotedblright=67, \textquotedblleft
horTh\textquotedblright=yes, \textquotedblleft menostat\textquotedblright%
=Post, \textquotedblleft pnodes\textquotedblright=6, \textquotedblleft
progrec\textquotedblright=150, \textquotedblleft tsize\textquotedblright=20,
\textquotedblleft tgrade\textquotedblright=1. Every picture in Fig.
\ref{f:gbsg-2_local_shape} shows how a separate feature contributes into the
prediction. Color vertical bars in every picture represent the normalized data
density for each feature. Only numerical features are considered for the local
explanation. It can be seen from Fig. \ref{f:gbsg-2_local_shape} that the
feature \textquotedblleft pnodes\textquotedblright\ (the number of positive
nodes) is the most important especially when its value is larger than $2$.
This conclusion follows from the fact that the corresponding shape function
significantly increases.%

%

%

\begin{figure}
[ptb]
\begin{center}
\includegraphics[
height=3.6858in,
width=3.6858in
]%
{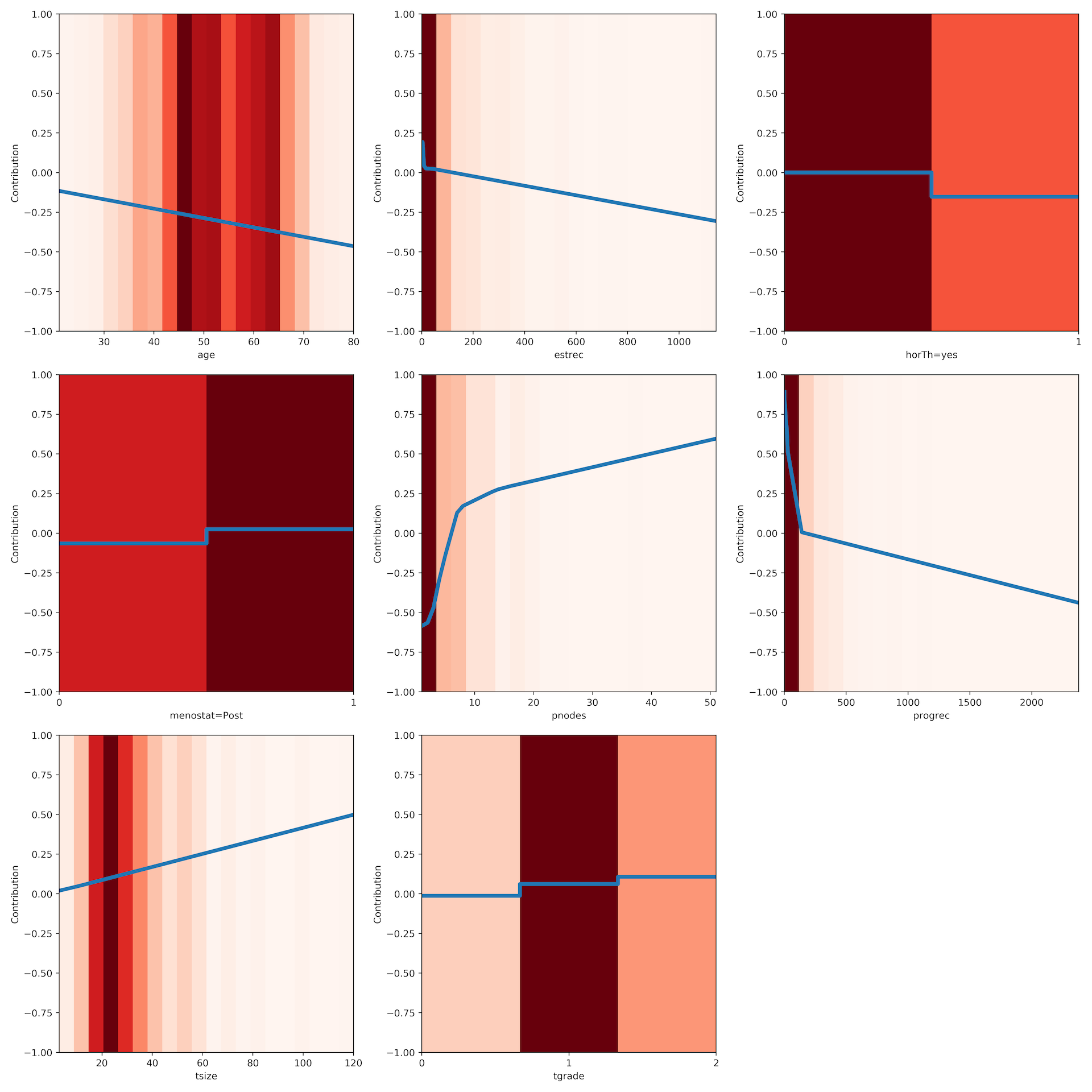}%
\caption{Shape functions for every feature obtained by means of SurvNAM for
the dataset GBSG2 and the global explanation}%
\label{f:gbsg-2_global_shape}%
\end{center}
\end{figure}

Let us consider how SurvNAM behaves for the case of the global explanation
when it is trained on the GBSG2 dataset. The shape functions for all features
are illustrated in Fig. \ref{f:gbsg-2_global_shape}. One can again see from
Fig. \ref{f:gbsg-2_global_shape} that the number of positive nodes
(\textquotedblleft pnodes\textquotedblright) is the most important feature. On
the one hand, features \textquotedblleft progrec\textquotedblright%
\ (progesterone receptor) and \textquotedblleft tsize\textquotedblright%
\ (tumor size) can be also regarded as important ones. On the other hand, if
we look at the normalized data density (color bar), then feature
\textquotedblleft age\textquotedblright\ becomes also important the main part
of its shape function is changed for a larger number of examples in
comparison, for example, with feature \textquotedblleft
progrec\textquotedblright. In a whole, one can observe from Figs.
\ref{f:gbsg-2_local_shape} and \ref{f:gbsg-2_global_shape} that tendency of
the feature value changes are similar for local and global explanations.

The RSF is characterized by the C-index equal to $0.676$. The C-index of
SurvNAM is $0.687$. The C-indices are computed on test data.

For comparison purposes, we also apply the well-known permutation feature
importance method \cite{Breiman-2001}. The largest importance values indicate
the top three features which are \textquotedblleft pnodes\textquotedblright,
\textquotedblleft progrec\textquotedblright, and \textquotedblleft
age\textquotedblright\ with the importance values $0.083$, $0.057$, $0.023$, respectively.%

\begin{figure}
[ptb]
\begin{center}
\includegraphics[
height=2.1041in,
width=4.0369in
]%
{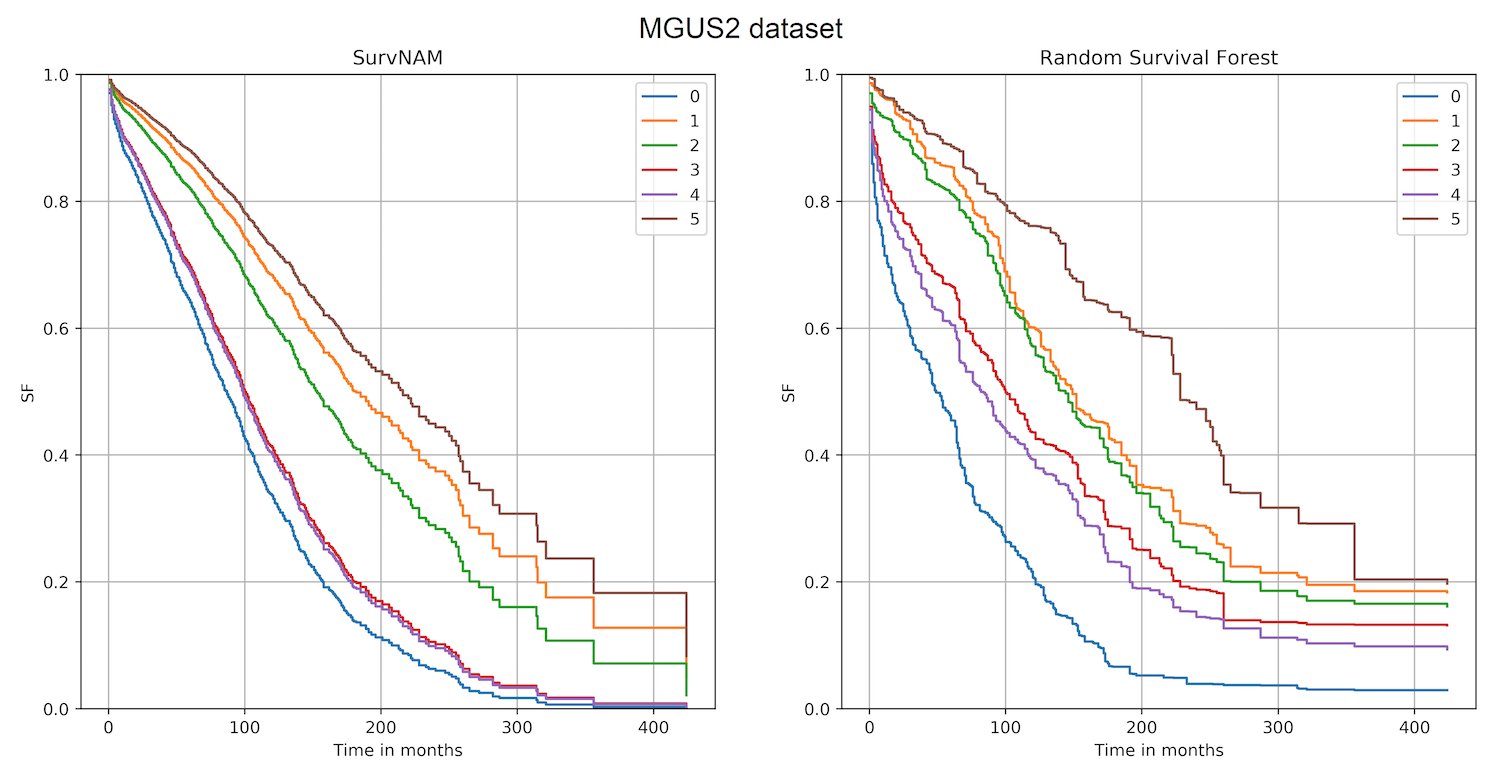}%
\caption{SFs obtained by means of SurvNAM (left picture) and the black-box RSF
model (right picture) for the dataset MGUS2 and the local explanation}%
\label{f:mgus2_local_sf}%
\end{center}
\end{figure}

\begin{figure}
[ptb]
\begin{center}
\includegraphics[
height=2.1283in,
width=4.0551in
]%
{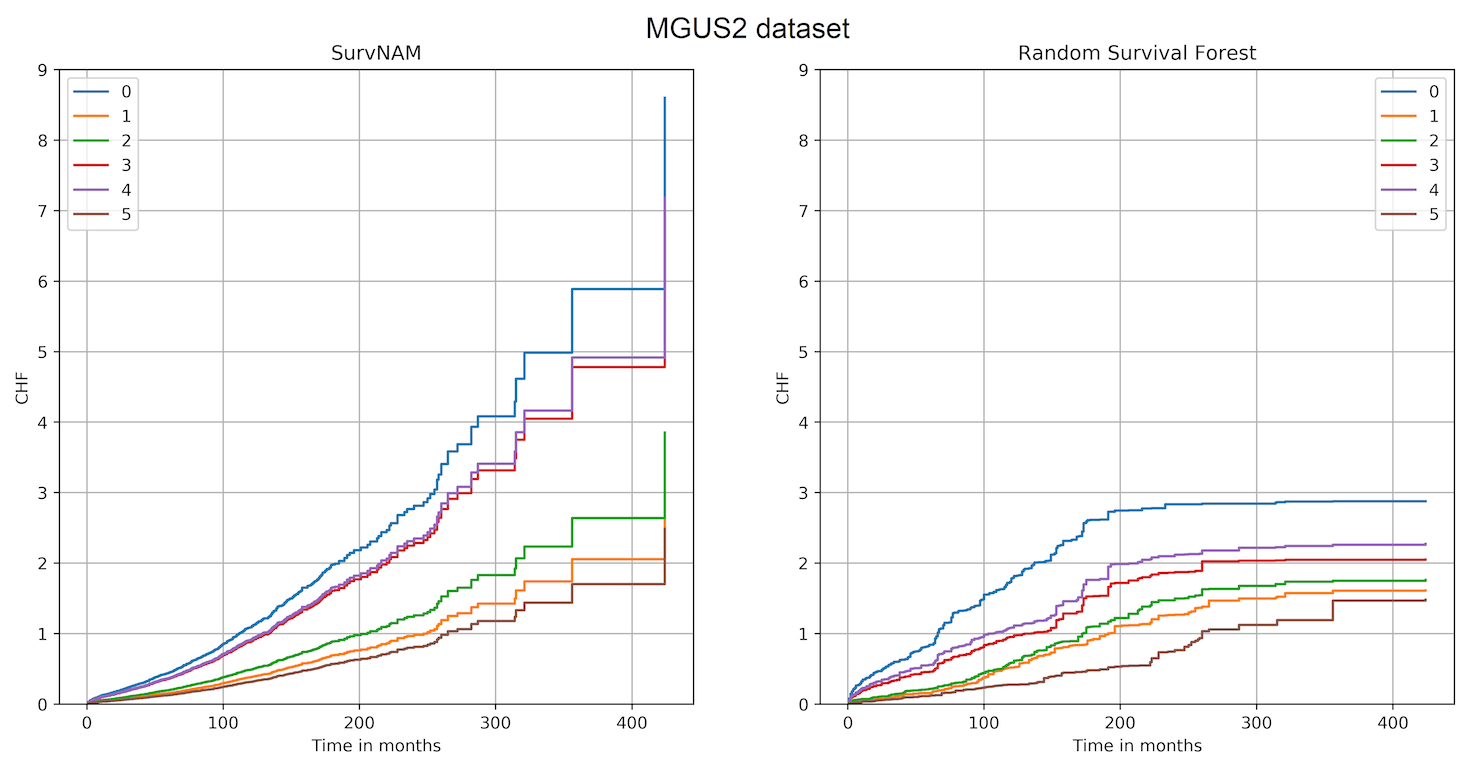}%
\caption{CHFs obtained by means of SurvNAM (left picture) and the black-box
RSF model (right picture) for the dataset MGUS2 and the local explanation}%
\label{f:mgus2_local_chf}%
\end{center}
\end{figure}

Another dataset for illustrating SurvNAM is the MGUS2 dataset. Figs.
\ref{f:mgus2_local_sf} and \ref{f:mgus2_local_chf} show SFs and CHFs obtained
by means of the RSF for $6$ randomly selected examples from the training set
(the right picture) and by means of the extended Cox model with the shape
functions $g_{i}(x_{i})$ of the GAM provided by SurvNAM (the left picture).
The pictures are similar to the corresponding pictures given in Figs.
\ref{f:gbsg-2_local_sf} and \ref{f:gbsg-2_local_chf}. It can be seen from
Figs. \ref{f:mgus2_local_sf} and \ref{f:mgus2_local_chf} that the SurvNAM
predictions are weakly approximate the RSF output. Two possible reasons can be
adduced to explain the above disagreement. First, the baseline CHF $H_{0}(t)$
may be unsatisfactory determined when the dataset is significantly
heterogeneous. Second, features are strongly correlated, and it is difficult
to construct the GAM for its using in the Cox model.

\begin{figure}
[ptb]
\begin{center}
\includegraphics[
height=2.7121in,
width=2.7121in
]%
{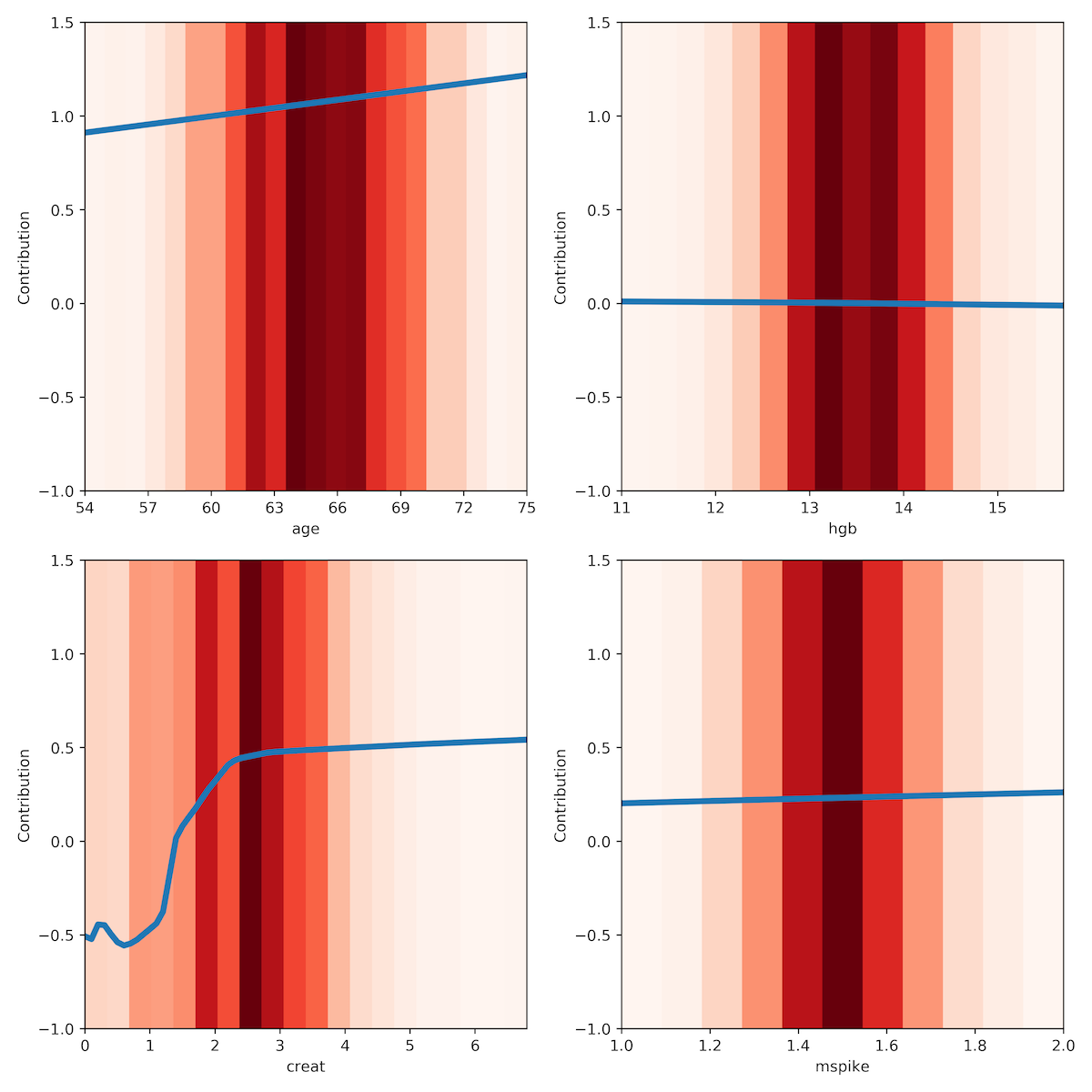}%
\caption{Shape functions for every feature obtained by means of SurvNAM for
the dataset MGUS2 and the local explanation}%
\label{f:mgus2_local_shape}%
\end{center}
\end{figure}

Shape functions for all features are illustrated in Fig.
\ref{f:mgus2_local_shape}. They are computed for an example with features
\textquotedblleft age\textquotedblright=65, \textquotedblleft
sex\textquotedblright=M, \textquotedblleft hgb\textquotedblright=13.5,
\textquotedblleft creat\textquotedblright=2.4, \textquotedblleft
mspike\textquotedblright=1.5. It can be seen from Fig.
\ref{f:mgus2_local_shape} that the feature \textquotedblleft
creat\textquotedblright\ (creatinine) is the most important. The feature
\textquotedblleft age\textquotedblright\ (age at diagnosis) can be also
regarded as important.%

%

%

\begin{figure}
[ptb]
\begin{center}
\includegraphics[
height=2.5746in,
width=3.8553in
]%
{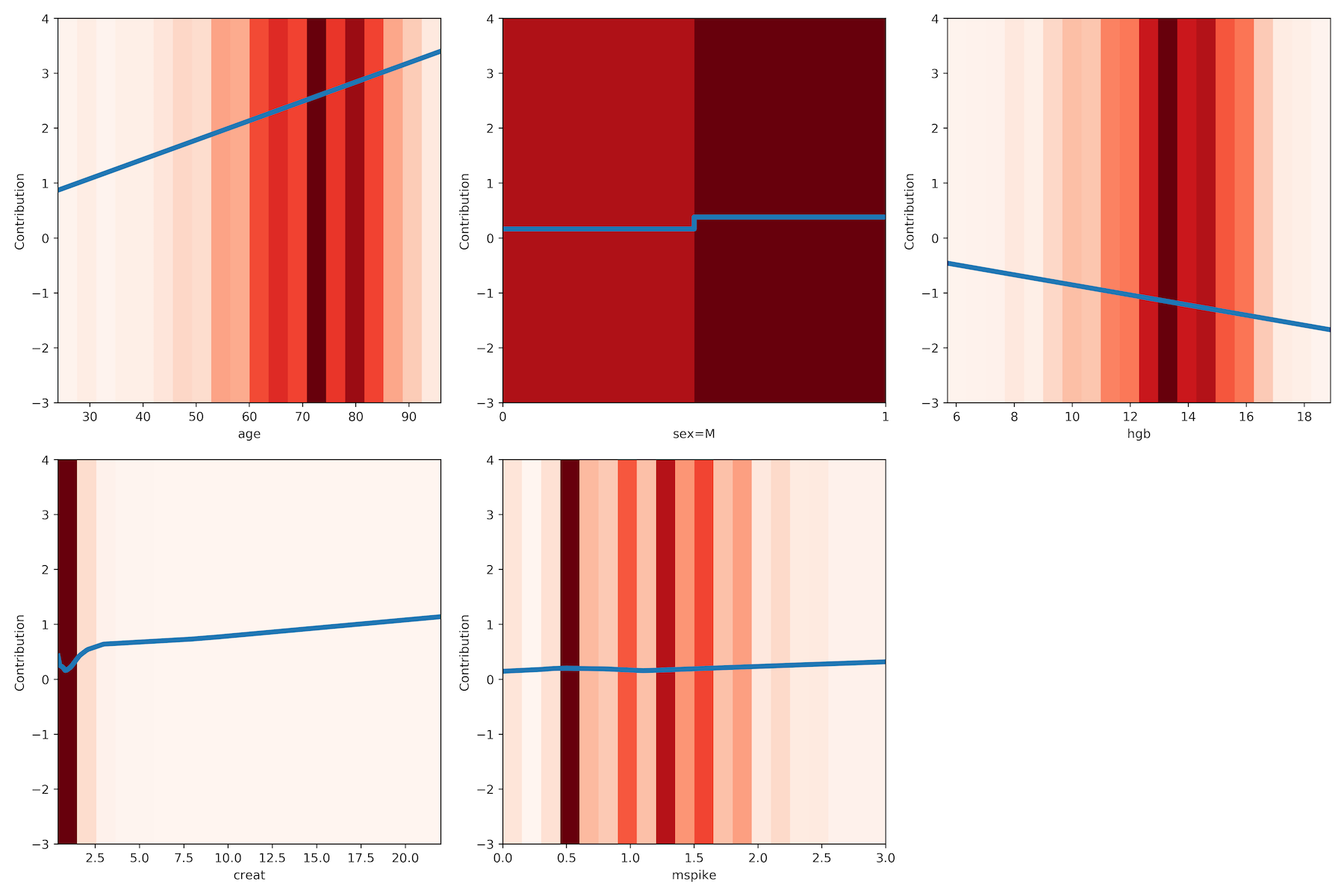}%
\caption{Shape functions for every feature obtained by means of SurvNAM for
the dataset MGUS2 and the global explanation}%
\label{f:mgus2_global_shape}%
\end{center}
\end{figure}

The shape functions for the global explanation are illustrated in Fig.
\ref{f:mgus2_global_shape}. It can be seen from Fig.
\ref{f:mgus2_global_shape} that \textquotedblleft age\textquotedblright\ and
\textquotedblleft hgb\textquotedblright\ are the most important features. This
conforms with results provided by the permutation feature importance method.
According to this method, the top three features are \textquotedblleft
age\textquotedblright, \textquotedblleft hgb\textquotedblright, and
\textquotedblleft creat\textquotedblright\ with the importance values $0.193$,
$0.041$, $0.027$, respectively.

The RSF is characterized by the C-index equal to $0.687$. The C-index of
SurvNAM is $0.691$. The C-indices are computed on test data.%

\begin{figure}
[ptb]
\begin{center}
\includegraphics[
height=2.0228in,
width=3.8917in
]%
{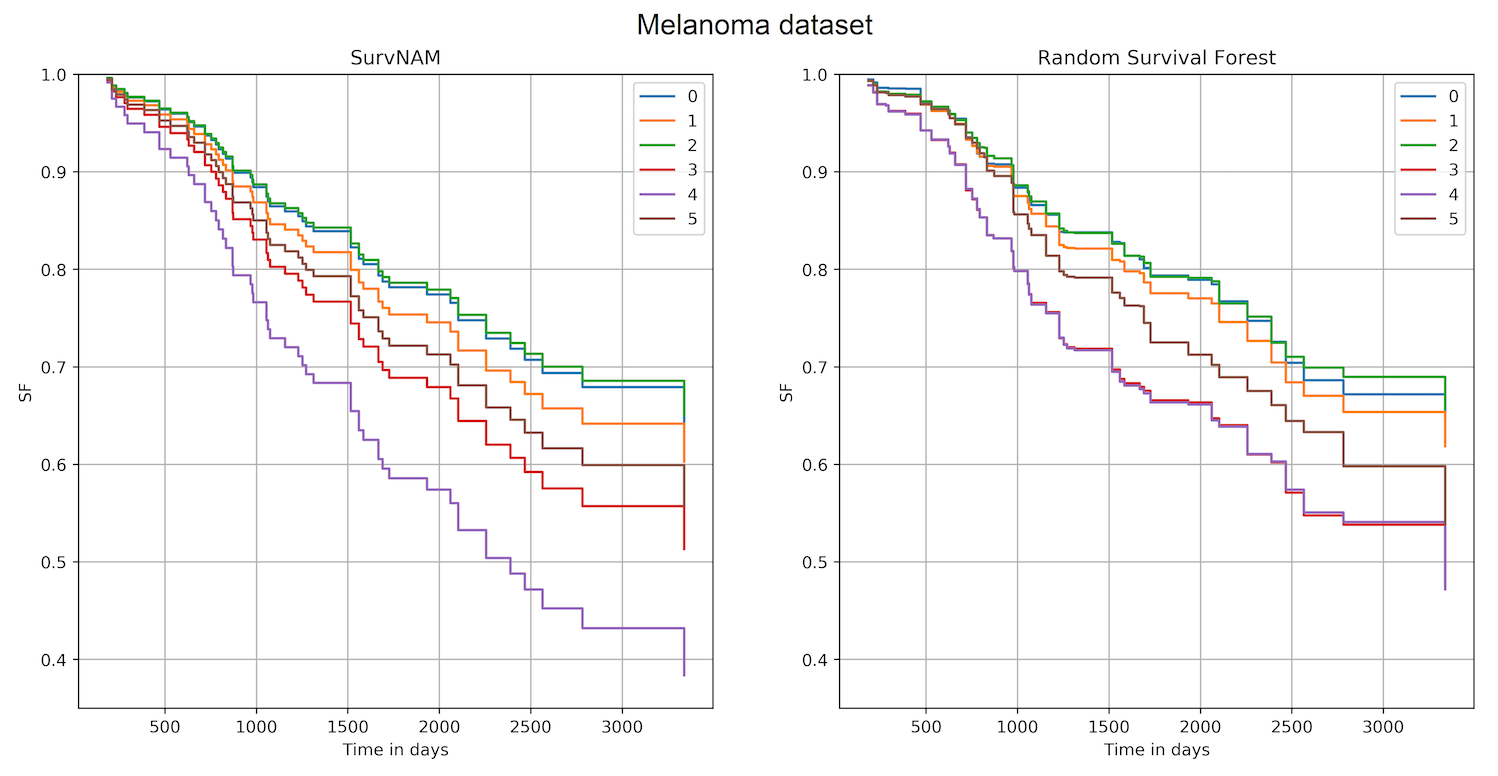}%
\caption{SFs obtained by means of SurvNAM (left picture) and the black-box RSF
model (right picture) for the dataset Melanoma and the local explanation}%
\label{f:melanoma_local_sf}%
\end{center}
\end{figure}

\begin{figure}
[ptb]
\begin{center}
\includegraphics[
height=2.0323in,
width=3.9159in
]%
{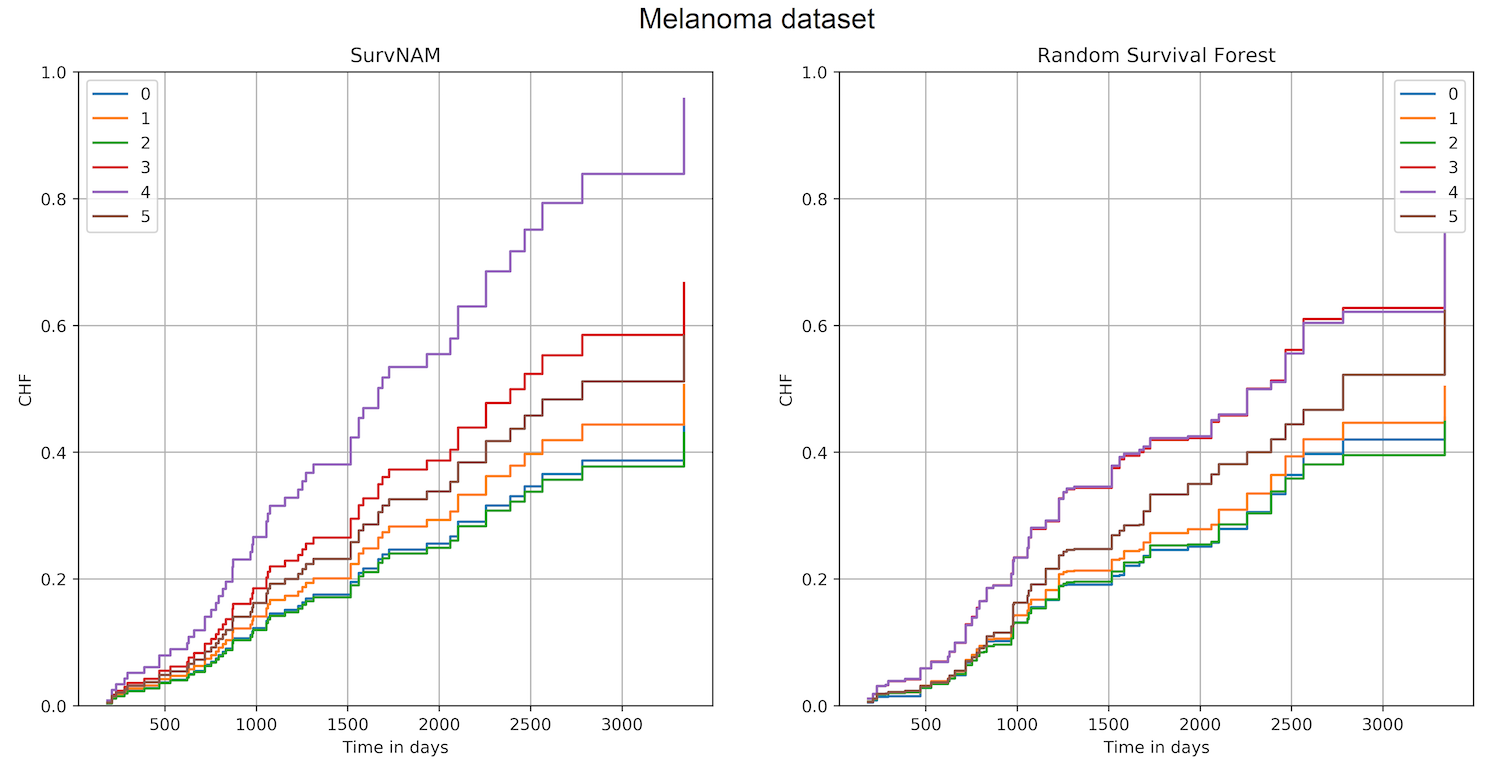}%
\caption{CHFs obtained by means of SurvNAM (left picture) and the black-box
RSF model (right picture) for the dataset Melanoma and the local explanation}%
\label{f:melanoma_local_chf}%
\end{center}
\end{figure}

\begin{figure}
[ptb]
\begin{center}
\includegraphics[
height=1.5082in,
width=3.007in
]%
{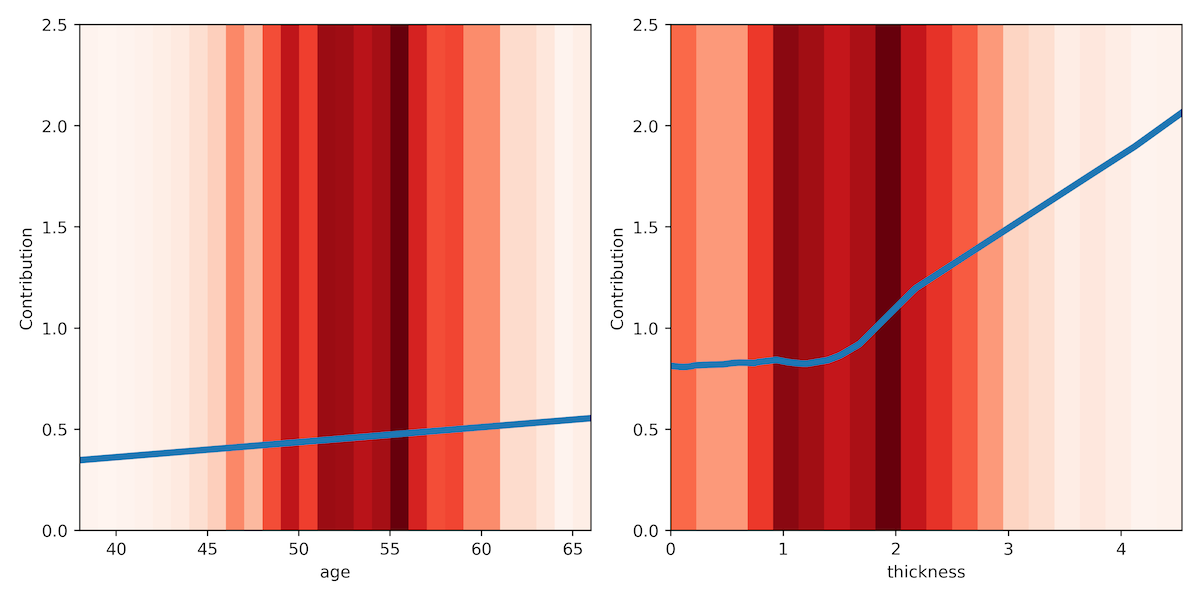}%
\caption{Shape functions for every feature obtained by means of SurvNAM for
the dataset Melanoma and the local explanation}%
\label{f:melanoma_local_shape}%
\end{center}
\end{figure}

Let us consider the Melanoma dataset. Figs. \ref{f:melanoma_local_sf} and
\ref{f:melanoma_local_chf} show SFs and CHFs obtained by means of the RSF for
$6$ randomly selected examples from the training set (the right picture) and
by means of the extended Cox model with the shape functions $g_{i}(x_{i})$ of
the GAM provided by SurvNAM (the left picture). In contrast to the MGUS2
dataset, SFs and CHFs provided by the RSF and SurvNAM are close to each other.
Shape functions for all features are illustrated in Fig.
\ref{f:melanoma_local_shape}. They are computed for an example with features
\textquotedblleft sex\textquotedblright=M, \textquotedblleft
age\textquotedblright=53, \textquotedblleft thickness\textquotedblright=1.62,
\textquotedblleft ulcer\textquotedblright=present. It can be seen from Fig.
\ref{f:melanoma_local_shape} that the feature \textquotedblleft
thickness\textquotedblright\ (the tumor thickness) is the most important.%

%

%

\begin{figure}
[ptb]
\begin{center}
\includegraphics[
height=2.9663in,
width=2.9663in
]%
{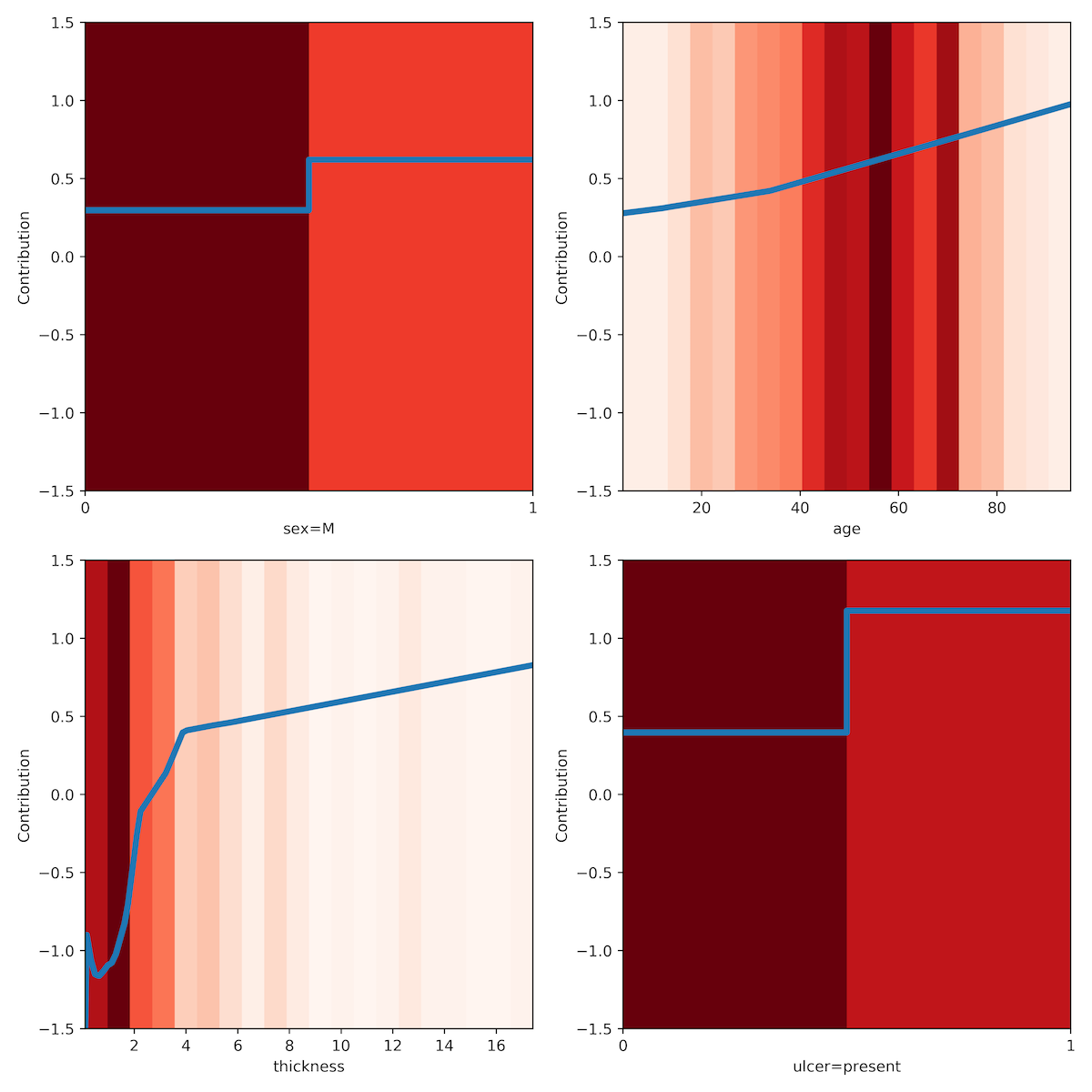}%
\caption{Shape functions for every feature obtained by means of SurvNAM for
the dataset Melanoma and the global explanation}%
\label{f:melanoma_global_shape}%
\end{center}
\end{figure}

The shape functions for the global explanation are illustrated in Fig.
\ref{f:melanoma_global_shape}. It can be seen from Fig.
\ref{f:melanoma_global_shape} that \textquotedblleft
thickness\textquotedblright\ and \textquotedblleft ulcer\textquotedblright%
\ are the most important features. The above result conforms with results
provided by the permutation feature importance method. According to this
method, the top three features are \textquotedblleft
thickness\textquotedblright, \textquotedblleft ulcer\textquotedblright, and
\textquotedblleft age\textquotedblright\ with the importance values $0.189$,
$0.033$, $0.026$, respectively. The importance values justify
\textquotedblleft thickness\textquotedblright\ and \textquotedblleft
ulcer\textquotedblright\ as the most important features. The RSF is
characterized by the C-index equal to $0.779$. The C-index of SurvNAM is
$0.788$. The C-indices are computed on test data.%

\begin{figure}
[ptb]
\begin{center}
\includegraphics[
height=1.9796in,
width=3.7853in
]%
{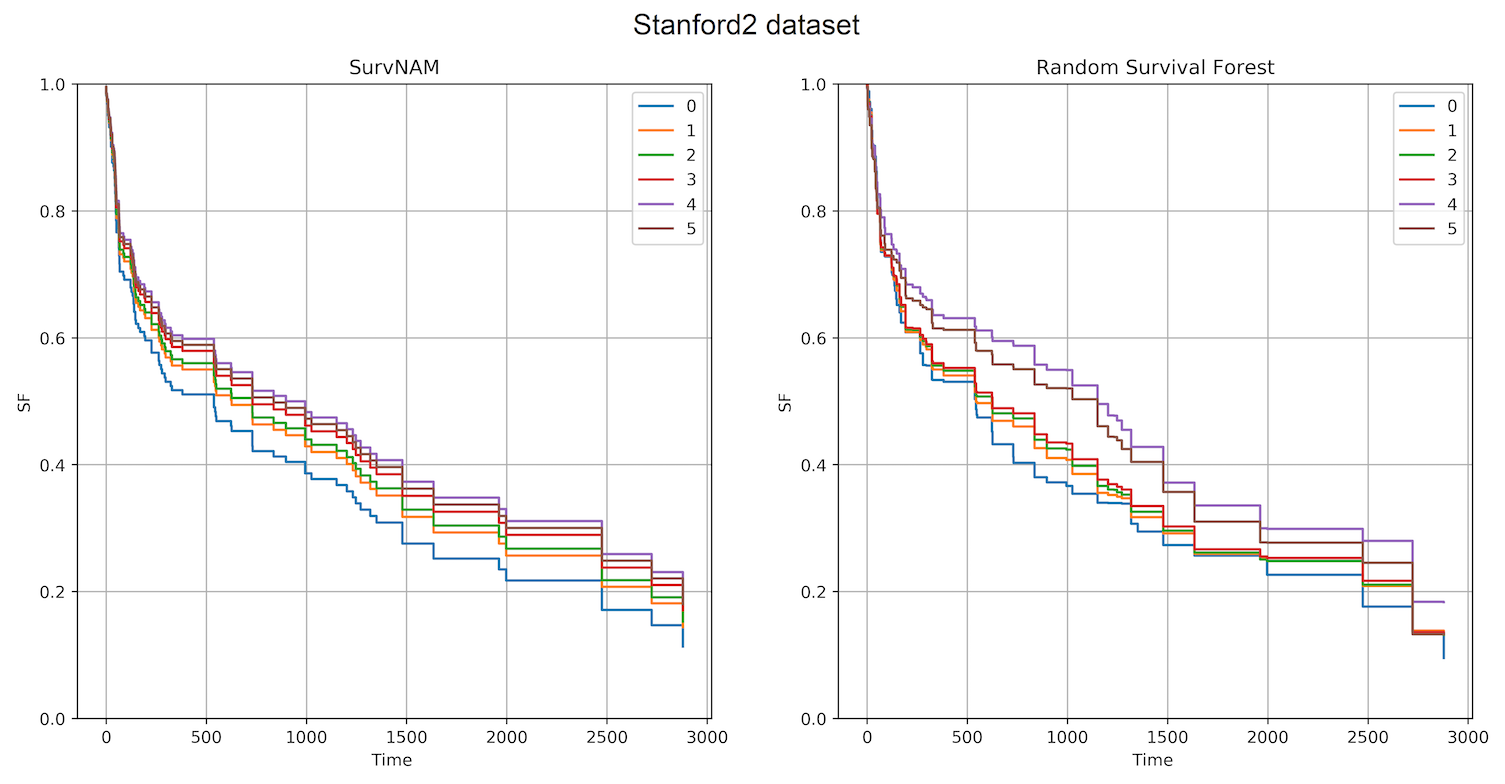}%
\caption{SFs obtained by means of SurvNAM (left picture) and the black-box RSF
model (right picture) for the dataset Stanford2 and the local explanation}%
\label{f:stanford2_local_sf}%
\end{center}
\end{figure}

\begin{figure}
[ptb]
\begin{center}
\includegraphics[
height=1.9821in,
width=3.87in
]%
{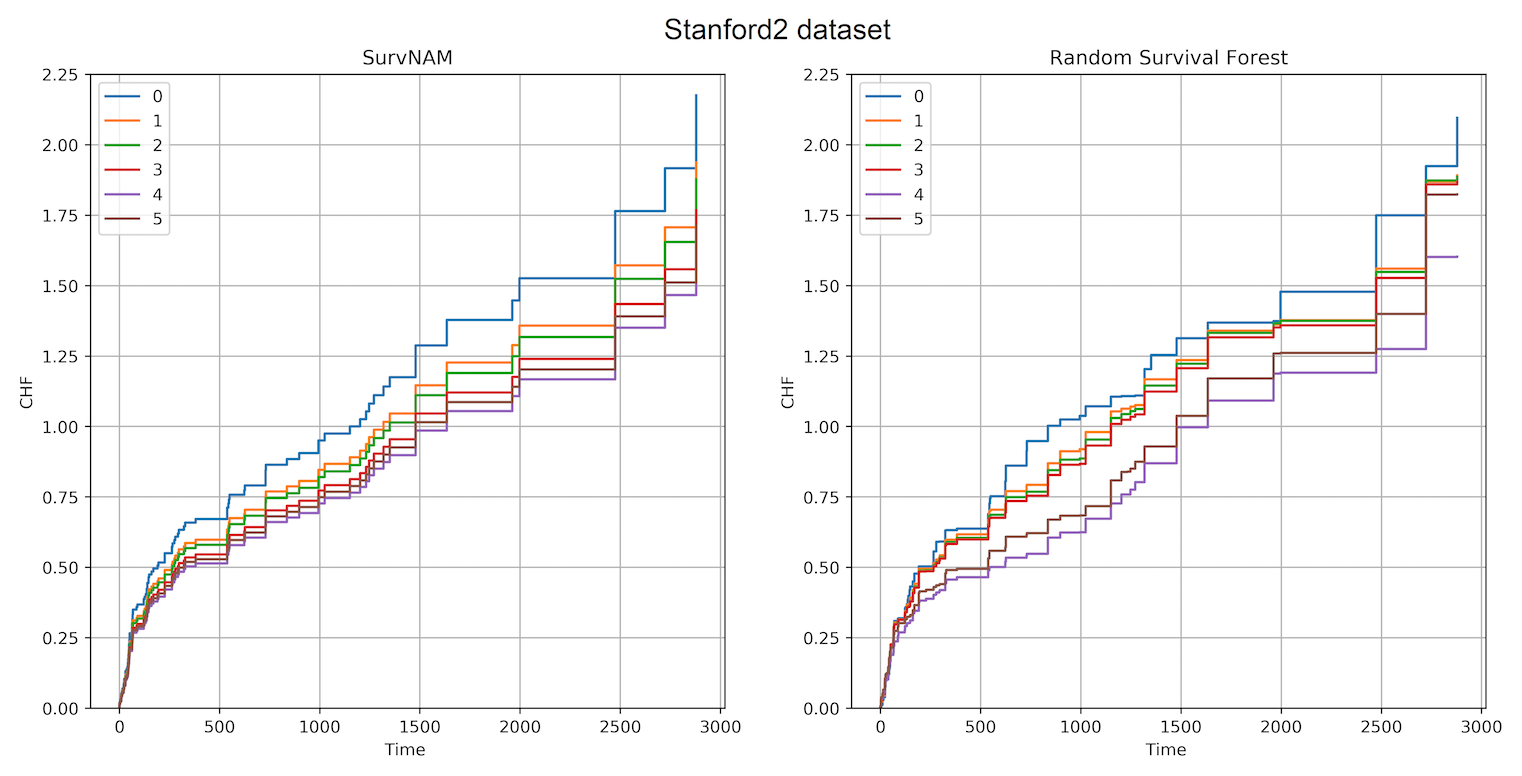}%
\caption{CHFs obtained by means of SurvNAM (left picture) and the black-box
RSF model (right picture) for the dataset Stanford2 and the local explanation}%
\label{f:stanford2_local_chf}%
\end{center}
\end{figure}

\begin{figure}
[ptb]
\begin{center}
\includegraphics[
height=1.5247in,
width=3.0372in
]%
{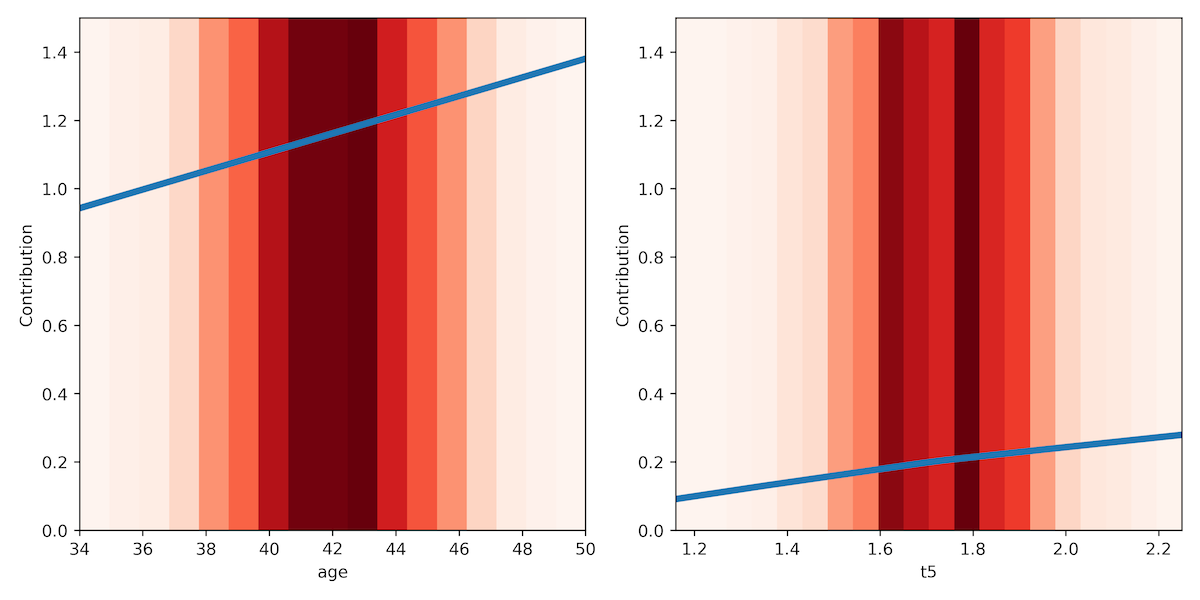}%
\caption{Shape functions for every feature obtained by means of SurvNAM for
the dataset Stanford2 and the local explanation}%
\label{f:stanford2_local_shape}%
\end{center}
\end{figure}

\begin{figure}
[ptb]
\begin{center}
\includegraphics[
height=1.5463in,
width=3.0848in
]%
{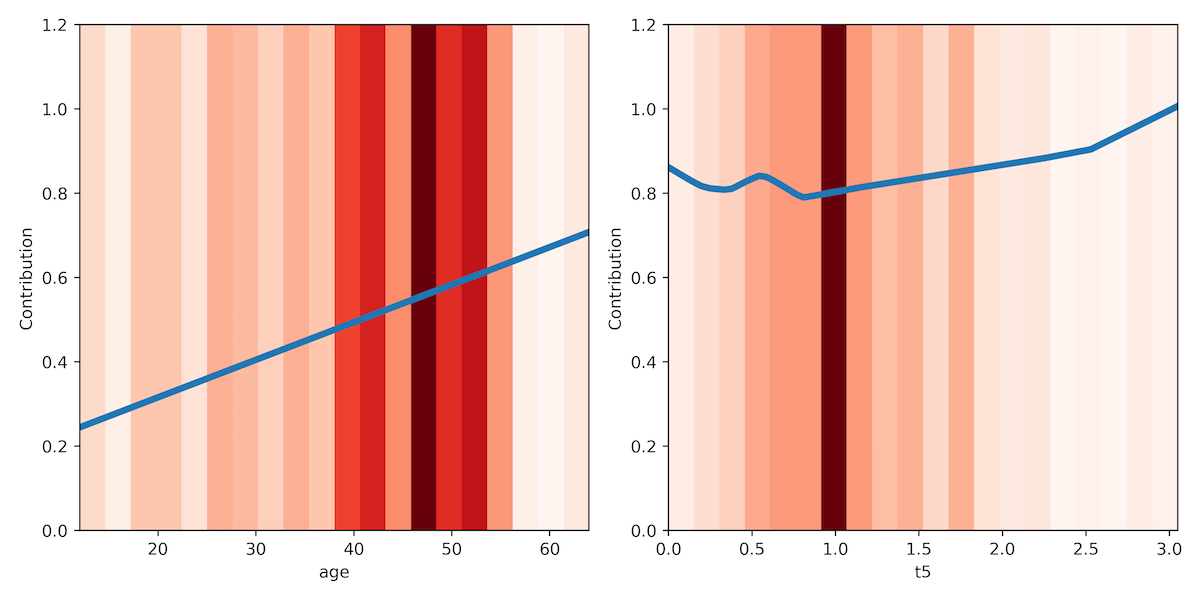}%
\caption{Shape functions for every feature obtained by means of SurvNAM for
the dataset Stanford2 and the global explanation}%
\label{f:stanford2_global_shape}%
\end{center}
\end{figure}

The next dataset is Stanford2. Figs. \ref{f:stanford2_local_sf} and
\ref{f:stanford2_local_chf} show SFs and CHFs obtained by means of the RSF for
$6$ randomly selected examples from the training set (the right picture) and
by means of the extended Cox model with the shape functions $g_{i}(x_{i})$ of
the GAM provided by SurvNAM (the left picture). In contrast to the Stanford2
dataset, SFs and CHFs provided by the RSF and SurvNAM are close to each other.
Shape functions for the local explanation are illustrated in Fig.
\ref{f:stanford2_local_shape}. They are computed for an example with features
\textquotedblleft age\textquotedblright=42, \textquotedblleft
t5\textquotedblright=1.74. It can be seen from Fig.
\ref{f:stanford2_local_shape} that the feature \textquotedblleft
age\textquotedblright\ is the most important. The same can be concluded for
the global explanation (see Fig. \ref{f:stanford2_global_shape}).

This result conforms with results provided by the permutation feature
importance method. According to this method, the important feature is
\textquotedblleft age\textquotedblright\ with the importance value $0.127$.
The importance value of \textquotedblleft t5\textquotedblright\ is $0.068$.
The RSF is characterized by the C-index equal to $0.643$. The C-index of
SurvNAM is $0.669$. The C-indices are computed on test data.%

%

%

%

\begin{figure}
[ptb]
\begin{center}
\includegraphics[
height=3.4679in,
width=3.4679in
]%
{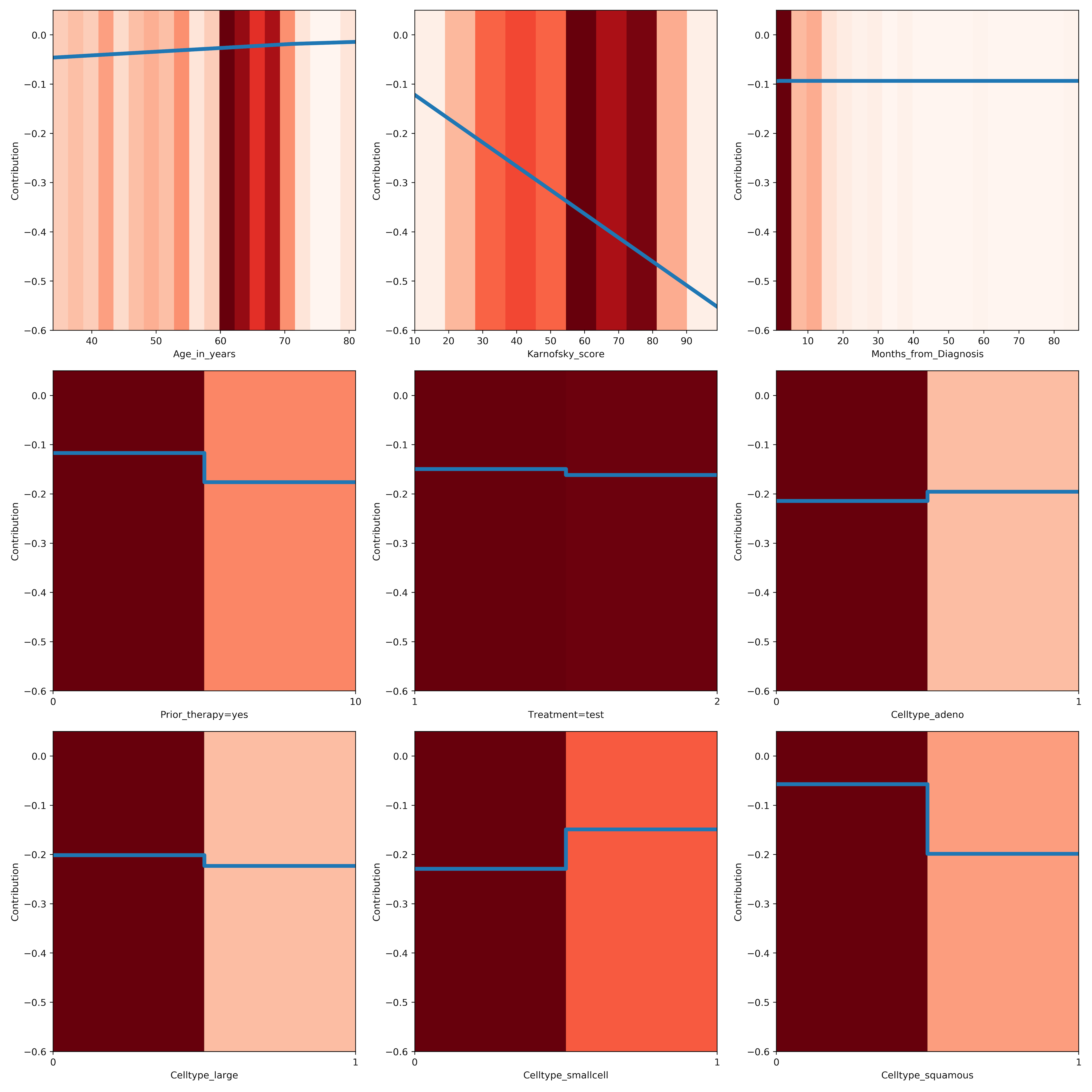}%
\caption{Shape functions for every feature obtained by means of SurvNAM for
the dataset Veteran and the global explanation}%
\label{f:veteran_global_shape}%
\end{center}
\end{figure}

Let us investigate SurvNAM by taking the well-known Veteran dataset. Examples
of the dataset Veteran consist of most categorical features. Therefore, we
consider only the global explanation for this dataset. The shape functions are
illustrated in Fig. \ref{f:veteran_global_shape}. It can be seen from Fig.
\ref{f:veteran_global_shape} that \textquotedblleft Karnofsky
score\textquotedblright\ and \textquotedblleft Celltype
squamous\textquotedblright\ are the most important features. According to the
permutation feature importance method, the top three features are
\textquotedblleft Karnofsky score\textquotedblright, \textquotedblleft
Celltype smallcell\textquotedblright, and \textquotedblleft Celltype
squamous\textquotedblright\ with the importance values $0.163$, $0.020$,
$0.008$, respectively. The above results do not totally conform with results
provided by SurvNAM. We can see from SurvNAM that \textquotedblleft Celltype
squamous\textquotedblright\ is more important in comparison with
\textquotedblleft Celltype smallcell\textquotedblright.

The RSF is characterized by the C-index equal to $0.725$. The C-index of
SurvNAM is $0.731$. The C-indices are computed on test data.%

\subsection{Numerical experiments with modifications of SurvNAM}

Let us consider how modifications of SurvNAM impact on the shape functions of features.

\begin{figure}
[ptb]
\begin{center}
\includegraphics[
height=1.6051in,
width=4.1122in
]%
{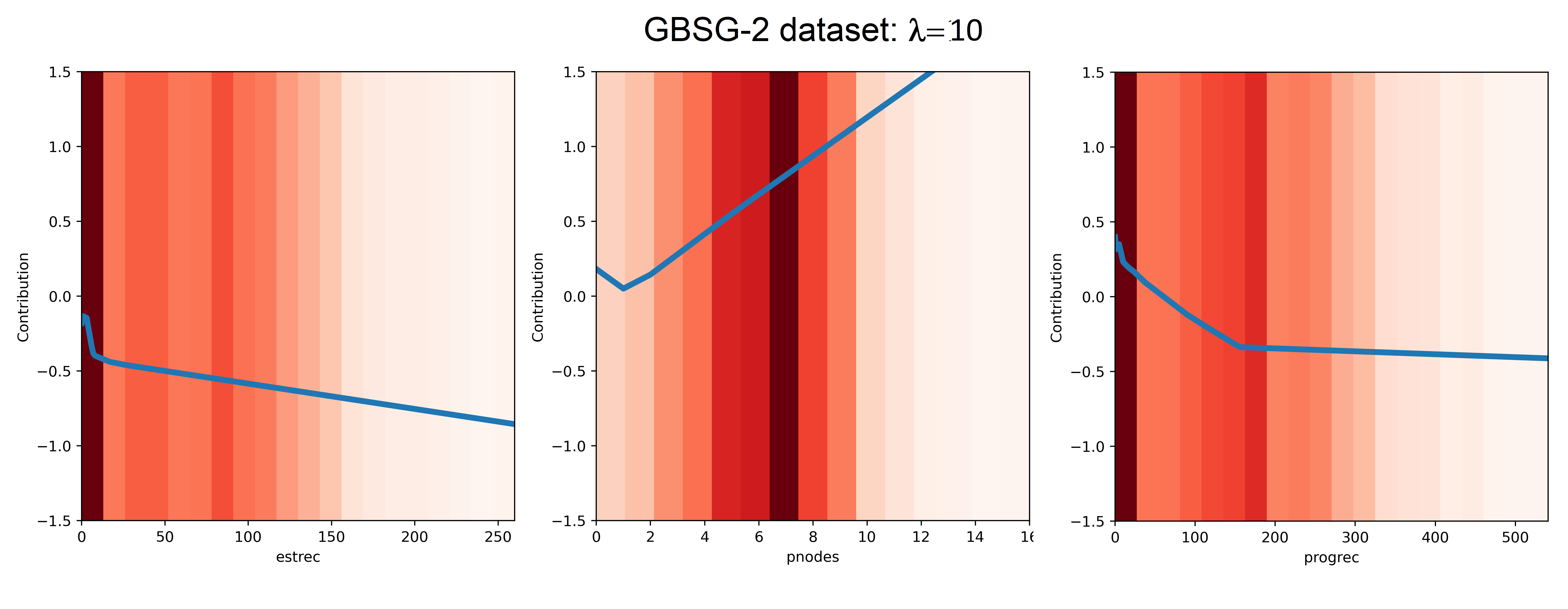}%
\caption{Shape functions for three important features obtained by means of the
first modification of SurvNAM for the dataset GBSG2 and the local explanation
by $\lambda=10$}%
\label{f:gbsg-2_local_shape_lambda_10}%
\end{center}
\end{figure}

First, we study the dataset GBSG2 for the local explanation. It can be seen
from Fig. \ref{f:gbsg-2_local_shape} that feature \textquotedblleft
pnodes\textquotedblright\ (the third picture in the first row) has an
unexplainable interval of decreasing when the feature values are in the
interval from $0$ to $2$. It is difficult to expect that the contribution to
the recurrence free survival time is reduced when the number of positive nodes
is $2$. One of the reasons of this behavior of the shape function is the
network overfitting. The same can be said about small values of features
\textquotedblleft estrec\textquotedblright\ (estrogen receptor) and
\textquotedblleft progrec\textquotedblright\ (progesterone receptor) (see Fig.
\ref{f:gbsg-2_local_shape}). Let us apply the first modification of SurvNAM
and train the network in accordance with the loss function (\ref{SurvNAM_51}).
The corresponding shape functions of analyzed features are depicted in Fig.
\ref{f:gbsg-2_local_shape_lambda_10}. Results in Fig.
\ref{f:gbsg-2_local_shape_lambda_10} are obtained under condition that
hyperparameter $\lambda$ controlling the strength of constraints for
parameters $\beta$ in (\ref{SurvNAM_51}) is $10$. It should be noted that
$\lambda=1$ does not impact on the shape functions. On the other hand, the
case $\lambda=100$ significantly reduces the approximation accuracy of the
extended Cox model and the black-box model. One can see from Fig.
\ref{f:gbsg-2_local_shape_lambda_10} that the first modification only partly
solves the network overfitting problem.%

\begin{figure}
[ptb]
\begin{center}
\includegraphics[
height=1.6734in,
width=1.5437in
]%
{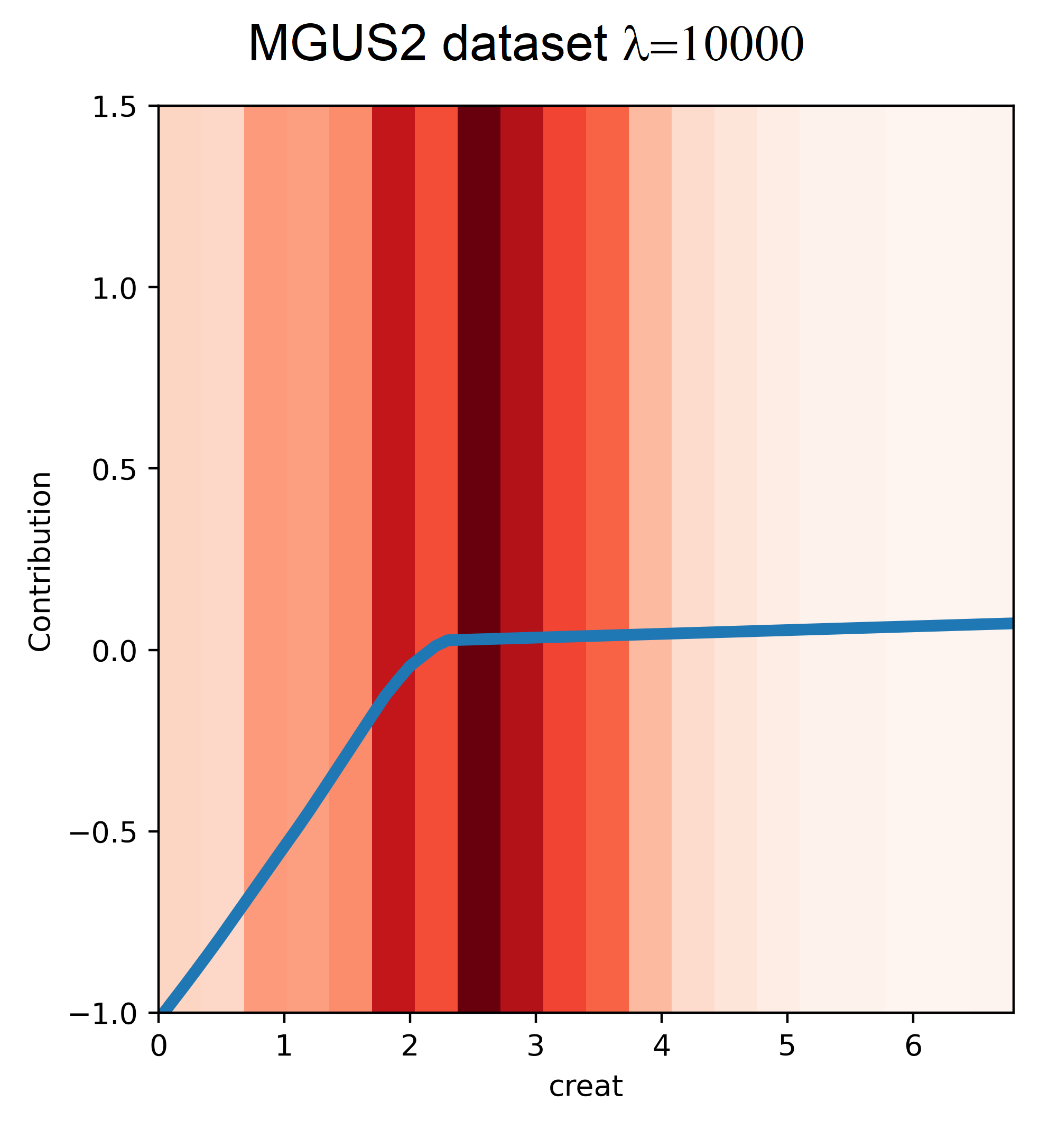}%
\caption{Shape functions for feature \textquotedblleft creat\textquotedblright%
\ obtained by means of the first modification of SurvNAM for the dataset MGUS2
and the local explanation by $\lambda=10000$}%
\label{f:mgus2_local_shape_lasso}%
\end{center}
\end{figure}

Let us consider now the dataset MGUS2 which has illustrated a strange behavior
of the shape function for the feature \textquotedblleft
creat\textquotedblright\ (creatinine) in Fig. \ref{f:mgus2_local_shape}. The
shape function of the feature obtained by using the first modification of
SurvNAM with the hyperparameter $\lambda=10000$ is depicted in Fig.
\ref{f:mgus2_local_shape_lasso}. Smaller values of $\lambda$ do not impact on
the form of the shape function. This implies that small values of parameters
$\beta_{k}$ are compensated by large values of functions $g_{k}$. This
undesirable property can be resolved by restricting values of functions
$g_{k}$, by introducing the shortcut connection, and using the $L_{2}$
regularization for the neural networks weights, i.e., by applying the second
modification of SurvNAM.%

\begin{figure}
[ptb]
\begin{center}
\includegraphics[
height=1.5645in,
width=3.9816in
]%
{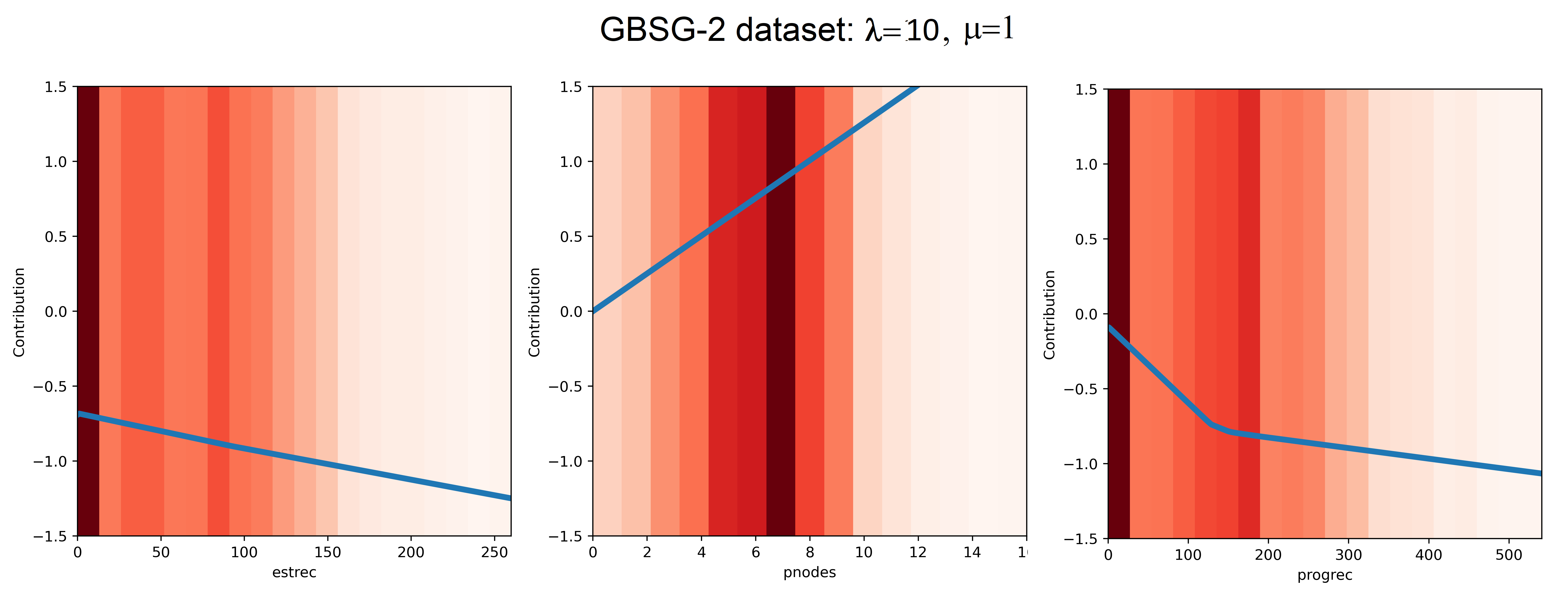}%
\caption{Shape functions for three important features obtained by means of the
second modification of SurvNAM with the shortcut connection for the dataset
GBSG2 and the local explanation by $\lambda=10$ and $\mu=1$}%
\label{f:gbsg2_local_shape_shortcut1}%
\end{center}
\end{figure}

The second modification of SurvNAM based on the shortcut connection is also
studied by analyzing three important features \textquotedblleft
estrec\textquotedblright, \textquotedblleft pnodes\textquotedblright,
\textquotedblleft progrec\textquotedblright\ of the dataset GBSG2, which show
possible overfitting of the neural network. The network is trained by using
the loss function (\ref{SurvNAM_53}) where hyperparameters $\lambda$ and $\mu$
are taken as $10$ and $1$, respectively. The corresponding shape functions are
shown in Fig. \ref{f:gbsg2_local_shape_shortcut1}. It can be seen from Fig.
\ref{f:gbsg2_local_shape_shortcut1} that artifacts, which take place in Fig.
\ref{f:gbsg-2_local_shape} as well as in Fig.
\ref{f:gbsg-2_local_shape_lambda_10}, are eliminated for all features. This
implies that the network is not overfitted. It is interesting to see how
values of $\alpha_{k}$ and $\omega_{k}$ allocated to features. Table
\ref{t:SurvNAM1} provides these values for numerical features. One can see
from Table \ref{t:SurvNAM1} that features \textquotedblleft
pnodes\textquotedblright\ and \textquotedblleft tsize\textquotedblright\ do
not use the non-linear part (functions $g_{k}$) and the corresponding shape
functions are represented as linear functions whereas features
\textquotedblleft estrec\textquotedblright\ and \textquotedblleft
progrec\textquotedblright\ cannot be represented by the linear functions. It
can clearly be seen for feature \textquotedblleft progrec\textquotedblright.
It is important to point out that restrictions on $\mathbf{W}$ in the form of
the regularization term $\left\Vert \mathbf{W}\right\Vert _{2}^{2}$ do not
allow functions $g_{k}$ to unboundedly increase.%

%

\begin{table}[tbp] \centering
\caption{Training parameters $\alpha _{k}$ and $\omega _{k}$ of the neural network trained on the dataset GBSG2 in accordance with the shortcut connection modification}%
\begin{tabular}
[c]{cccccc}\cline{3-6}%
Features & age & estrec & pnodes & progrec & tsize\\\hline
$\alpha_{k}$ & $0.315$ & $0.775$ & $0$ & $0.929$ & $0$\\\hline
$1-\alpha_{k}$ & $0.685$ & $0.225$ & $1.000$ & $0.071$ & $1$\\\hline
$\omega_{k}$ & $0.006$ & $0.315$ & $0.315$ & $0.315$ & $0.315$\\\hline
$(1-\alpha_{k})\omega_{k}$ & $0.004$ & $0.071$ & $0.315$ & $0.022$ &
$0.315$\\\hline
\end{tabular}
\label{t:SurvNAM1}%
\end{table}%

\begin{figure}
[ptb]
\begin{center}
\includegraphics[
height=1.5826in,
width=3.7057in
]%
{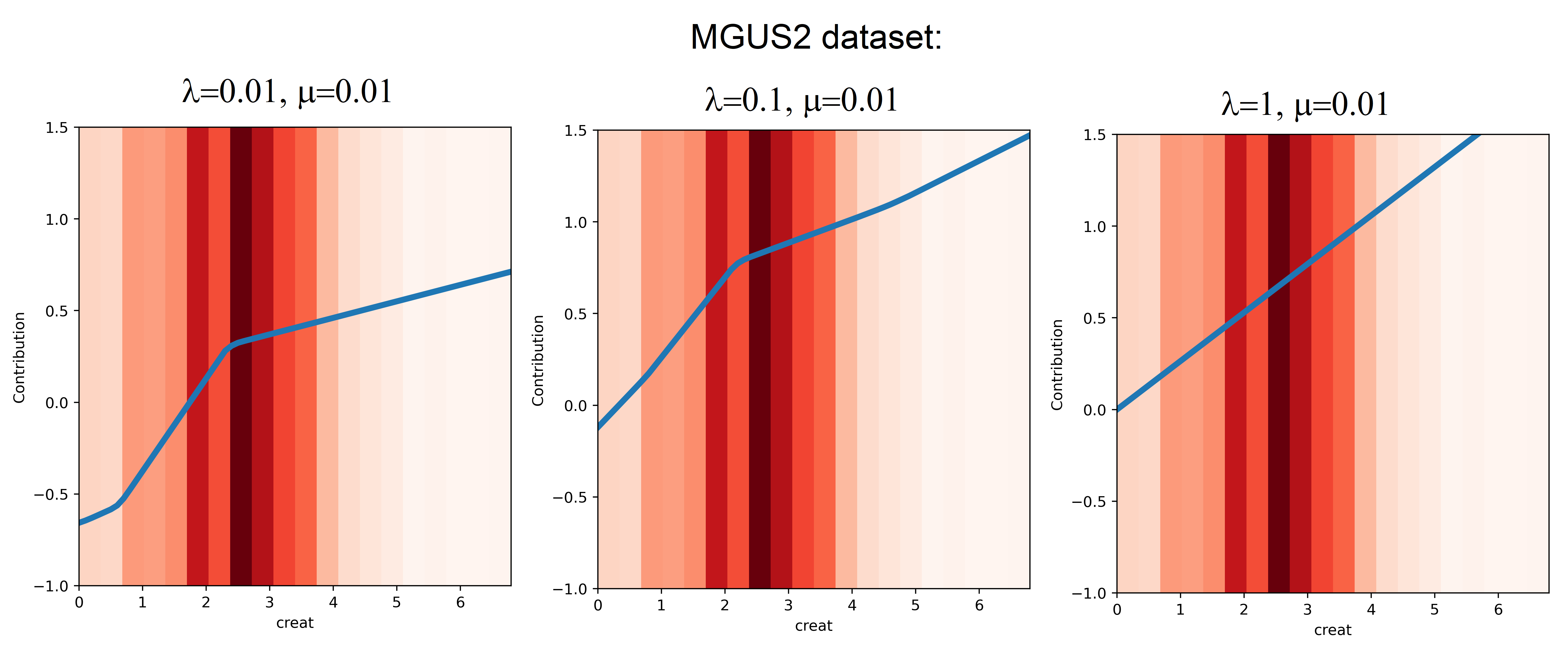}%
\caption{Shape functions for the feature \textquotedblleft
creat\textquotedblright\ obtained by means of the second modification of
SurvNAM with the shortcut connection for the dataset MGUS2 and the local
explanation by different values of $\lambda$ and $\mu=0.01$}%
\label{f:mgus2_local_shape_shortcut1}%
\end{center}
\end{figure}

Similar results are obtained for the dataset MGUS2. Shape functions for the
feature \textquotedblleft creat\textquotedblright\ obtained by means of the
second modification of SurvNAM with the shortcut connection by three values of
$\lambda=0.01$, $0.1$, $1$ and $\mu=0.01$ are shown in Fig.
\ref{f:mgus2_local_shape_shortcut1}. It can be seen from Fig.
\ref{f:mgus2_local_shape_shortcut1} how the linear part of the loss function
begins to dominate with increase of the hyperparameter $\lambda$. The
corresponding parameters $\alpha_{k}$, $\omega_{k}$ and their combinations by
$\lambda=0.01$ and $\lambda=1$ are represented in Table \ref{t:SurvNAM2}. One
can compare values of parameters for different strengths of the
regularization. It is interesting to note that $\alpha_{k}$ for the
\textquotedblleft bad\textquotedblright\ feature \textquotedblleft
creat\textquotedblright\ changes from $0.922$ by $\lambda=0.01$ till $0$ by
$\lambda=1$. This implies that linear part of the loss function totally
dominates when $\lambda=1$. This fact is confirmed by the corresponding
pictures in Fig. \ref{f:mgus2_local_shape_shortcut1}.%

%

\begin{table}[tbp] \centering
\caption{Training parameters $\alpha _{k}$ and $\omega _{k}$ of the neural network trained on the dataset MGUS2 in accordance with the shortcut connection modification}%
\begin{tabular}
[c]{ccccccccc}\cline{3-9}
& \multicolumn{4}{c}{$\lambda=0.01$} & \multicolumn{4}{c}{$\lambda=1$%
}\\\cline{3-9}%
Features & age & hgb & creat & mspike & age & hgb & creat & mspike\\\hline
$\alpha_{k}$ & $0.976$ & $0.995$ & $0.922$ & $0.820$ & $0.788$ & $0.262$ & $0$
& $0$\\\hline
$1-\alpha_{k}$ & $0.024$ & $0.005$ & $0.078$ & $0.180$ & $0.212$ & $0.738$ &
$1$ & $1$\\\hline
$\omega_{k}$ & $0.830$ & $0.765$ & $1.095$ & $1.012$ & $0.861$ & $0.768$ &
$0.264$ & $0.489$\\\hline
$(1-\alpha_{k})\omega_{k}$ & $0.020$ & $0.004$ & $0.085$ & $0.182$ & $0.183$ &
$0.567$ & $0.264$ & $0.489$\\\hline
\end{tabular}
\label{t:SurvNAM2}%
\end{table}%

\section{SurvNAM vs. SurvLIME}

An explanation method called SurvLIME has been proposed by Kovalev et al.
\cite{Kovalev-Utkin-Kasimov-20a}. This method also explains predictions of the
black-box survival models. Therefore, one of the important questions is how
SurvNAM and SurvLIME relate with each other.

First, they solve the same task of the survival model explanation. However,
SurvLIME solves only a task of the local explanation whereas SurvNAM can solve
the local as well as global explanations. As a result, SurvLIME uses only
perturbation technique whereas SurvNAM may use only the training set if the
global explanation is aimed.

Second, both the methods are based on the Cox model approximation. This is the
basic idea behind the methods. However, SurvLIME is based on using the
original Cox model with linear relationship between covariates whereas SurvNAM
uses an extension of the Cox model for approximation. According to the
extension, the sum of some functions of covariates is used instead of their
simple linear sum. As a result, the extended Cox model may significantly
improve its approximation to the explained model.

Third, SurvLIME solves a complex optimization problem for computing
coefficients of covariates in the Cox model whereas the neural network is
trained in SurvNAM. In fact, the neural network also implicitly solves the
convex optimization problem. However, it is more flexible from the point of
view of various optimization variants. In particular, various types of the
regularization can be added to SurvNAM whereas SurvLIME is restricted only by
regularizations which remain the optimization problem in the classes of linear
and quadratic programming problems.

Fourth, SurvLIME provides the explanation results in the form of coefficients
reflecting the feature impacts on predictions whereas SurvNAM provides the
feature contribution functions which indirectly show the feature impact. One
can see from the shape functions how the contribution of each feature changes
with change of its values.

It can be seen from the above that SurvNAM and SurvLIME have many common
elements, but these elements have the different content. It is difficult to
evaluate which method is better. Every method has advantages and
disadvantages. For example, on the one hand, SurvLIME solves the quadratic
optimization problem which has a unique solution and can be exactly solves by
several available algorithms. SurvNAM requires training the neural network
which can be overfitted or just can provide inaccurate predictions. On the
other hand, the neural network is a more flexible tool and can provide
direction for various extensions and modifications of SurvNAM. On the one
hand, the neural network can deal with very complex objects, for example, with
images. Therefore, SurvNAM has the important advantage in comparison with
SurvLIME. On the other hand, it is difficult to construct and to train the
neural network when the number of features is very large in SurvNAM whereas
SurvLIME can deal with the large number of features because the number of
features weakly impact on the optimization problem complexity. Though we have
to point out that the number of features impact on the perturbation procedure.

We see again that both the methods are comparable. Nevertheless, SurvNAM is
more flexible, its results are more informative, it has great potentialities
for extending and for solving more complex survival problems than SurvLIME.

\section{Conclusion}

Among various distances between the predictions of the black-box model and the
extended Cox model applied to constructing the loss function, we have
considered only the Euclidean distance. However, it is also interesting to
study other metrics which preserve the loss function convexity, for example,
the Manhattan distance (the $L_{1}$-norm) and the Chebyshev distance (the
$L_{\infty}$-norm). These distances may provide unexpected results whose
studying is a direction for further research.

There are two main difficulties of the proposed method. First, it represents
results in the form of the feature shape functions as measures of the feature
importance. However, explanations are simpler if they are represented by a
single number measuring the feature importance. Second, the method uses neural
networks which may be overfitted and require many examples for their training.
One of the approaches overcoming the above difficulties is the ensemble of
gradient boosting machines \cite{Konstantinov-Utkin-21}. An extension of this
approach to the survival model explanation problem may be an interesting
direction for further research.

Another interesting problem is to consider how combinations of correlated
features impact on the survival model prediction. A direct way is to train
$2^{m}$ neural subnetworks such that each pair of features is fed to each
subnetwork. This way requires to train $2^{m}$ subnetworks. Therefore, another
approach should be developed to take into account correlated features. This is
a direction for further research.

It is interesting to investigate how different regularization terms impact on
results. We have illustrated how the well-known Lasso method with the
corresponding regularization term can be applied to removing unimportant shape
functions in the extended Cox model and improves the results. However, other
regularization methods can be of the large interest. It should be also noted
that the shortcut connection trick used in the proposed modifications of
SurvNAM can be also applied to other tasks different from the survival model
explanation. We have not also studied how time-dependent features might change
the proposed approach. The above questions can be viewed as directions for
further research.

The proposed method is efficient mainly for tabular data. However, it can be
also adapted to the image processing which has some inherent peculiarities.
The adaptation is another interesting direction for research in future.

\section*{Acknowledgement}

This work is supported by the Russian Science Foundation under grant 21-11-00116.

\bibliographystyle{unsrt}
\bibliography{Boosting,Classif_bib,Deep_Forest,Explain,Explain_med,Lasso,MYBIB,MYUSE,Robots,Survival_analysis}

\end{document}